\documentclass[journal]{IEEEtran}
\usepackage[utf8]{inputenc}
\usepackage{newunicodechar}
\usepackage[colorlinks,linkcolor=red]{hyperref}
\usepackage{amsmath}
\usepackage{graphicx}
\usepackage{booktabs} 
\usepackage{amssymb}
\usepackage{float}
\usepackage{comment}
\usepackage{mathrsfs}
\usepackage{color}
\usepackage{xcolor}
\usepackage{subfig}
\usepackage{subcaption}
\usepackage{multirow} 
\usepackage{booktabs}
\usepackage{multirow}
\usepackage{makecell}
\usepackage{threeparttable}
\usepackage{array}
\usepackage{siunitx}
\usepackage{adjustbox}
\usepackage[dvipsnames, table]{xcolor}
\usepackage{caption}
\usepackage{cite}
\usepackage{textcomp}
\usepackage{tcolorbox}
\usepackage{mdframed} 

\newcommand{\MetricHead}{&mIoU(\%)$\uparrow$ & $\mathrm{P_d}(\%)\uparrow$ & $\mathrm{F_a}(10\textsuperscript{-6})\downarrow$ & $\mathrm{AUC}(\%)\uparrow$}

\definecolor{MyMagenta}{RGB}{208, 48, 142}

\definecolor{headbg}{gray}{0.92} 
\definecolor{headbg1}{rgb}{0.85, 0.90, 0.95}
\ifCLASSINFOpdf
\else
\fi
\begin{document}
\title{STGBD-Net: Spatio-temporal Gradient Basis Decomposition Network for Infrared Small Target Detection}
%
%
%

\author{Chen Hu, ~\IEEEmembership{Graduate Student Member,~IEEE},
        Mingyu Zhou,
        Shuai Yuan,
        Hongbo Hu,
        Zhenming Peng,~\IEEEmembership{Senior Member,~IEEE},
        Tian Pu\textsuperscript{*},
        and Xiying Li\textsuperscript{*},~\IEEEmembership{Member,~IEEE}

\thanks{Manuscript received XXX XXX, XXX; revised XXX XXX, XXX.}

\thanks{\textit{(Corresponding authors: Xiying Li; Tian Pu.)}}
\thanks{Chen Hu and Xiying Li are with the School of Intelligent Systems Engineering, Sun Yat-sen University, Shenzhen, China (e-mail: huch67@mail2.sysu.edu.cn; stslxy@mail.sysu.edu.cn)
}
\thanks{Mingyu Zhou, Hongbo Hu, Zhenming Peng, and Tian Pu are with the School of Information and Communication Engineering and the Laboratory of Imaging Detection and Intelligent Perception, University of Electronic Science and Technology of China, Chengdu 610054, China (e-mail:  2024010904031@std.uestc.edu.cn, 202111012015@std.uestc.edu.cn, zmpeng@uestc.edu.cn, putian@uestc.edu.cn). 
}
\thanks{
Shuai Yuan is with the School of Instrument Science and Opto-Electronics Engineering, Hefei University of Technology, Hefei, China. (shuaiyuan@hfut.edu.cn)
}
\thanks{This work was supported by the National Natural Science Foundation of China (Grant No.U21B2090).}
}

\markboth{Journal of \LaTeX\ Class Files,~Vol.~14, No.~8, August~2015}%
{Shell \MakeLowercase{\textit{et al.}}: Bare Demo of IEEEtran.cls for IEEE Journals}
%

\maketitle

\begin{abstract}
A key challenge in infrared small target detection (IRSTD) is that weak target signal responses are easily obscured by strong background clutter, frequently resulting in missed detections. 
While traditional gradient-based methods attempt to capture fine details, their robustness is limited by the static fusion of multi-directional gradient features.
In this paper, we rethink feature fusion from the perspective of Basis Decomposition Theory and propose a novel framework that reformulates the process into an explicit and adaptive decomposition-and-reconstruction paradigm.
Specifically, we introduce the Basis Decomposition Module (BDM) and its specialized variant, the Gradient Decomposition Module (GDM) for IRSTD. 
GDMs treat the normalized gradient features as basis vectors to reconstruct a new feature, thereby maintaining detailed structures and highlighting infrared small targets.
By integrating GDMs into a lightweight three-stage U-Net, we develop two unified architectures: the Spatial Gradient Basis Decomposition Network for single-frame detection and the Spatio-temporal Gradient Basis Decomposition Network for multi-frame scenarios.
Extensive experiments demonstrate that our networks achieve state-of-the-art (SOTA) performance across multiple benchmarks, offering a superior balance between detection accuracy and computational efficiency.
Our codes will be made public at: \url{https://github.com/greekinRoma/IRSTD_HC_Platform}.
\end{abstract}


\section{Introduction}
\label{sec:intro}
\IEEEPARstart{U}{nlike} visible light imaging, infrared sensing maintains reliable performance in adverse weather conditions, offering high detection reliability, concealment, and mobility.
Infrared Search and Track (IRST), a primary application of infrared sensing, is crucial for maritime surveillance \cite{Zhang2024IRPruneDet, Zhu2025PFF}, early warning systems \cite{iMoPKL2025, MICPL2025}, and unmanned aerial vehicle detection \cite{yuan2025ASCNet}.

Infrared small target detection (IRSTD) is regarded as a crucial task in IRST systems, attracting growing research interest. 
Nevertheless, IRSTD remains more challenging than many other detection tasks due to dim targets and complex backgrounds \cite{Zhu2024GSTUnet}.
Consequently, robust IRSTD remains a fundamental yet challenging computer vision task \cite{dai2025SeRankDet, zhang2025daaf, zhang2025coupled, yuan2025EDGSP}.

Numerous model-based approaches have been proposed for single-frame infrared small target detection (SIRSTD). 
Representative methods include tensor-based, contrast-based and filter-based techniques \cite{kong2022LogTFNN, cao2021infrared, dai2017reweighted, wang2019miss}. 
Although these methods can achieve strong performance under certain conditions, they often struggle in complex scenes and require careful tuning of hyperparameters, which limits their robustness and practical applicability. 

Deep learning–based approaches for infrared small target detection (IRSTD) can effectively overcome challenges caused by dynamic and complex backgrounds. 
Data-driven methods have recently dominated IRSTD by overcoming the limitations of hyperparameter-sensitive traditional approaches \cite{Zhang2025IRmamba, li2025hstnet, li2025mmidnet, zhu2025shifting}.
According to their applications, these methods could be categorized into two main groups: spatial models and spatio-temporal models for SIRSTD and multi-frame infrared small target detection (MIRSTD).

\begin{figure}
    \centering
    \subfloat[CDC method with fixed coefficients]{%
        \includegraphics[width=\linewidth]{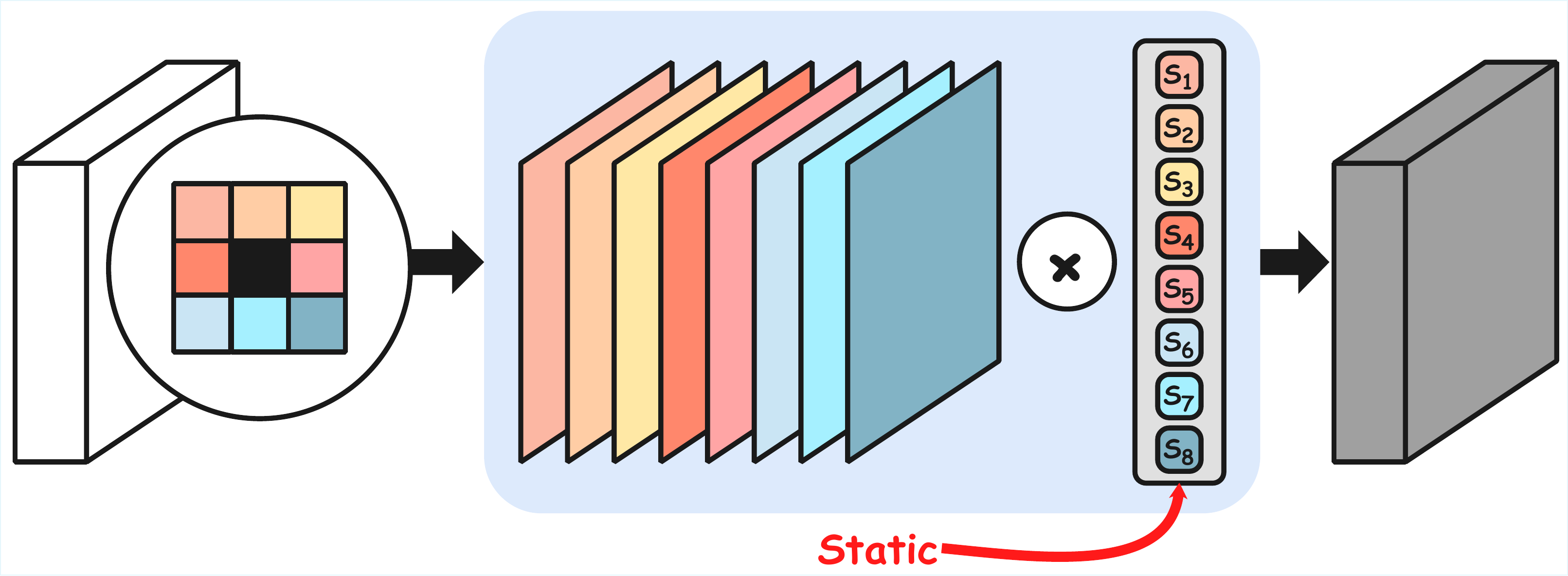}%
    }%
    \hfill%
    \subfloat[Frame-difference method with fixed coefficients]{%
        \includegraphics[width=\linewidth]{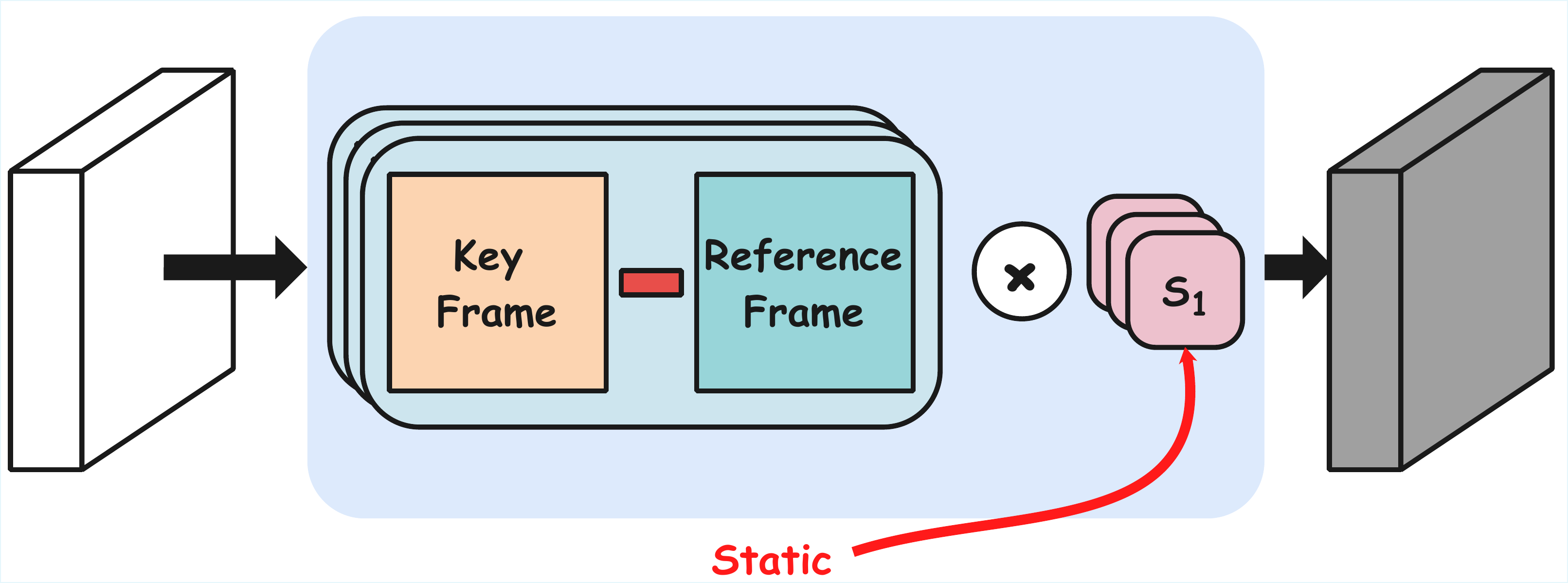}%
    }%
    \caption{Standard CDC and frame-difference operations, both of which employ predetermined static coefficients to combine gradient information. (a) Difference kernels are employed to extract spatial gradient features across various directions, which are then combined using static weights. (b) Temporal gradient features are extracted by subtracting the key frame from reference frames, followed by fusion using static coefficients.}
    \label{fig:coefficient_comparison}
\end{figure}

Algorithms for SIRSTD are mainly applied in scenarios where temporal information is unavailable or unreliable, such as low-frame-rate imaging, severe platform motion, dynamic backgrounds, and strict real-time requirements \cite{zhao2022SIRSTsurvey}.
There are many methods introduced for SIRSTD \cite{dai2025SeRankDet, DaiWuZhouBarnard2021Asymmetric, WuHongChanussot2023UIU-Net}, achieving competitive performance.
 
Methods designed for MIRSTD are well suited to scenarios where the imaging platform remains stable, background variations are mild, and targets exhibit consistent and predictable motion patterns across consecutive frames \cite{zhao2022SIRSTsurvey}.
Recently, with the release of infrared multi-frame datasets \cite{Li2025dtum, SunBaiYangBai2023Receptive-Field}, MIRSTD-based methods utilize motion information to address scenarios in which true targets and false alarms exhibit similar structural characteristics \cite{zhang2025mocid}. 

Gradient-based approaches play a crucial role in both SIRSTD and MIRSTD tasks \cite{dai2021attentional, Li2025dtum}.
Therefore, central difference convolutions (CDCs) \cite{Wu2025Salm} and difference-frame methods \cite{Yan2023STDMANet}, as the representative gradient-based approaches, have attracted increasing attention.
These approaches emphasize details by computing and fusing gradients along spatial or temporal dimensions \cite{Zhang2021ECA}.
However, they suffer from a fundamental limitation due to the use of static coefficients, as the importance of different directions varies across images, as illustrated in Fig. \ref{fig:coefficient_comparison}.

Basis decomposition can be regarded as a suitable approach for assigning dynamic coefficients to varied gradient responses, providing stronger interpretability and lower computational burden compared to the Self-Attention Mechanism. 
From the perspective of basis decomposition, any input feature vector can be expressed as a linear combination of a set of basis vectors (or basis features) with corresponding coefficients. 
These coefficients are obtained by projecting the input onto the basis features and can be regarded as intrinsic characteristics of the input vector, as they are uniquely determined once the input is fixed.

Building on this Basis Decomposition Theory, we design a unified and lightweight module for IRSTD, referred to as the Basis Decomposition Module (BDM).
In contrast to original basis decomposition, we only select the task-related basis vectors (or basis features) to reconstruct enhanced feature representations, thereby highlighting the features we aim to emphasize.

Based on the BDM, we develop Gradient Decomposition Modules (GDMs) for IRSTD. 
Each GDM treats gradient responses as basis features to reconstruct the input, which effectively enhances the representation of dim targets.

To minimize the computational overhead of the model incorporating GDM, we strategically compress the channel dimensions within the backbone. However, such a lightweight design can lead to a loss in feature representational capacity. To compensate for this, we adopt a three-stage U-Net as our backbone. This multi-stage architecture allows for iterative feature refinement and stronger aggregation, recovering the performance lost from channel compression \cite{Liu2025RRCANet}.

By integrating GDMs into this three-stage backbone, we propose two task-specific networks: the Spatial Gradient
Basis Decomposition Network (SGBD-Net) for SIRSTD and the Spatio-temporal Gradient
Basis Decomposition Network (STGBD-Net) for MIRSTD. 
Both networks achieve robust performance while maintaining a very small parameter count.
The main contributions of this work are summarized as follows:
\begin{enumerate}
    \item 
    We design a general BDM based on the theory of basis decomposition and derive GDMs from this BDM for IRSTD, enhancing task-relevant features while suppressing irrelevant information.
    \item 
    We introduce the GDMs to a three-stage U-Net and propose the SGBD-Net for SIRSTD, to enhance dim and small target features with a small number of parameters.
    \item 
    We further build the STGBD-Net, which is based on the SGBD-Net and effectively highlights moving targets while suppressing the static clutter for MIRSTD.
\end{enumerate}
\section{Related Work}

\label{sec:related}

\subsection{Single-frame Infrared Small Target Detection}
Existing SIRSTD methods can be categorized into two methodological paradigms: model-driven methods and data-driven methods. 

\textbf{Model-driven methods} are generally divided into three main categories: filter-based, human-visual-system-inspired, and optimization-based approaches.
Filter-based methods \cite{Qin2019Facetkernel} enhance targets while suppressing background clutter using handcrafted filters.
Human-visual-system-inspired methods \cite{WeiYouLi2016Multiscale, Han2021WLCM} exploit perceptual characteristics of the human visual system to highlight targets.
Optimization-based methods \cite{gao2013infrared, ZhangPeng2019Infrared} formulate IRSTD as an optimization problem, detecting targets by separating them from the background.
Despite their strong interpretability and computational efficiency, these methods are sensitive to parameter selection and tend to be unstable in complex, cluttered scenes.

\textbf{Data-driven methods} could overcome the challenges of the model-driven methods. 
Dai et al. have firstly proposed the Asymmetric Contextual Modulation (ACM) \cite{DaiWuZhouBarnard2021Asymmetric} and Attentional Local Contrast Network (ALCNet) \cite{dai2021attentional} for SIRSTD.
Dense Nested Attention Network (DNANet) \cite{RenLiHanShu2021DNANet} supports adaptive interactions between feature layers to prevent the loss of targets. 
Furthermore, UIUNet \cite{WuHongChanussot2023UIU-Net} employs a simple and effective “U-Net in U-Net" framework to enhance both global and local contrast information. 
Receptive-field and Direction-induced Attention Network (RDIAN) \cite{SunBaiYangBai2023Receptive-Field} utilizes the characteristics of targets to solve the imbalance between targets and background. 
Infrared Small-target Detection U-Net (ISTDU-Net) \cite{HouZhangTanXiZhengLi2022ISTDU-Net} introduces a fully connected layer in the skip connection to suppress the backgrounds with similar structures from the global receptive field. 
Attention-guided Pyramid Context Network (AGPCNet) \cite{ZhangCaoPuPeng2021AGPCNet} employs an attention-guided context block, providing the model with a perspective on both inner and global patches.
Spatial-Channel Cross Transformer Network (SCTransNet) \cite{Yuan2024SCtransNet} utilizes the Transformer to extract practical global information. 
Yuan et al. \cite{YUAN20261SP-KAN} were the first to introduce the KAN model to the field of SIRSTD, achieving remarkable performance.
Compared with conventional model-driven networks, the models exhibit strong robustness.
Zhang et al. \cite{fc3net2022mingjinzhang} propose FC3-Net, which consists of a Fine-detail guided Multi-level Feature Compensation (F-MFC) module and a Cross-level Feature Correlation (CFC) module to improve model performance.
Furthermore, they incorporate textual information to guide the network \cite{SAIST2025zhang}.
Nevertheless, they suffer from higher computational complexity, which is critical in real-time applications.
Based on the above discussion, efficiency should be regarded as one of the most important merits.

SIRSTD methods can be applied to a wide range of scenarios, as they remain effective regardless of whether the acquired images form a sequence. 
When image sequences are available, incorporating temporal information can further improve detection performance. Accordingly, some MIRSTD models are designed to explicitly leverage temporal features.

\subsection{Multi-frame Infrared Small Target Detection}
Some MIRSTD methods integrate spatial and temporal modeling into a unified framework, while others decouple them into two separate modules.
Thus, we could categorize MIRSTD approaches into two groups: unified and separate models.

\textbf{The unified models} simultaneously extract spatial and temporal information.
Sliced Spatio-temporal Network (SSTNet) \cite{Chen2024SSTNet} utilizes the ConvLSTM node to extract the motion information of the small infrared targets. 
The Lightweight Asymmetric Spatial Feature Network (LASNet) \cite{Chen2024LASNet} utilizes motion feature extraction and motion-affinity fusion for extracting temporal information. 
The Triple-domain Strategy (Tridos) \cite{Duan2024Triple} employs frequency-aware enhancement to enhance the detection of dim infrared targets.
The Local Motion Aware Transformer (LMAFormer) \cite{huang2024lmaformer} utilizes local motion-aware attention to enhance the weak target motion information in MIRSTD.
DQAligner \cite{Deng2026DQAligner} introduces novel techniques for better tracking and detecting small moving infrared targets by improving motion discrimination, maintaining temporal consistency, and enabling more flexible target localization across frames.
These end-to-end architectures can effectively integrate spatial and temporal cues, while they typically require heavy computation and high memory usage, limiting their applications in real-time infrared tracking systems. Furthermore, these approaches lack intuitive interpretability, compared with separation-based methods.

Generally, \textbf{separation-style methods} decouple spatial and temporal modeling into two dedicated modules. According to how temporal information is extracted, these methods can be further categorized into two groups: pre-extraction methods, which extract temporal information before spatial modeling, and post-extraction methods, which perform temporal extraction after spatial feature extraction.

\textbf{Pre-extraction models} extract motion features prior to the SIRSTD network.
Most existing methods obtain temporal information through model-driven approaches, such as Energy Accumulation (IFEA) \cite{Du2022IFEA} and the Spatio-temporal Differential Multiscale Attention Network (STDMANet) \cite{Yan2023STDMANet}.
The Recurrent Feature Refinement (RFR) model \cite{ying2025rfr} captures long-term temporal dependencies by recursively refining the outputs of an SIRSTD network over time.
These approaches enhance target energy by computing gradients between the current frame and reference frames as a pre-processing step before network input.

\textbf{Post-extraction models} obtain temporal information from the outputs of single-frame neural networks.
The Direction-code Temporal U-shape Module (DTUM) \cite{Li2025dtum} extracts motion features by further modeling temporal information based on the results of an SIRSTD network.
In these models, temporal modules can be easily plugged into different single-frame SIRSTD networks, enabling efficient temporal modeling with low additional computational cost.

For both SIRSTD and MIRSTD, gradient operations across spatial and temporal domains are fundamental. Accordingly, a summary of gradient-based methods is necessary.

\subsection{Gradient-based Methods}
Gradient-based Operators in IRSTD. Recently, gradient-based operators have gained significant attention in IRSTD \cite{Su2025Rapid} due to their inherent ability to highlight fine-grained variations.

In the spatial domain, standard convolutions often struggle to capture the subtle local contrasts that define sub-pixel targets against complex backgrounds. To mitigate this, several methods adapt classical descriptors, such as Local Binary Patterns (LBP) \cite{Ahonen2006LBP} and Gabor filters \cite{JAIN1991Gaborfilter}, to explicitly model local intensity transitions \cite{Xu2017LBPConv}. A pivotal advancement is the Central Difference Convolution (CDC) \cite{yu2020CDC}, which enhances detail representation by incorporating gradient-level information directly into the kernel. Building on this, DEANet \cite{Chen2024DEANet} employs a parallel architecture to extract complementary semantic and detailed features. For more irregular geometries, the Deformable Kernel Network \cite{Kim2021Deformable} adaptively reconfigures sampling positions to model complex spatial variations.

In the temporal domain, inter-frame gradients serve as a fundamental cue for motion modeling \cite{Yan2023STDMANet}. For instance, 3D Central Difference Convolution (3D-CDC) \cite{GENG20253DCDC} extends the gradient operator into the temporal dimension, effectively suppressing static background clutter through motion-aware filtering.

Existing gradient-based methods rely on static convolution weights, which limits their flexibility under varying conditions. Shifting away from conventional approaches, we propose a novel architecture that treats information extraction as a decomposition and reconstruction process. 
By selectively utilizing specific features for reconstruction, we can enhance the task-related features. 
In the context of IRSTD, we choose gradient information to effectively enhance and distinguish small targets from complex backgrounds.

To sum up, GDM is the modified BDM for IRSTD, relying on the gradient information as the basis features.

\section{Methodology}
\label{sec:method}
\subsection{Preliminaries}
\subsubsection{The procedure of basis decomposition}
Given an input feature vector $\vec{x}\in \mathbb{R}^{c_{\text{full}}}$, where $c_{\text{full}}$ denotes the full dimensionality, $\vec{x}$ can be represented or reconstructed through an orthonormal basis $\mathbf{B} = [\vec{b}_1, \vec{b}_2, \dots, \vec{b}_{c_{\text{full}}}] \in \mathbb{R}^{c_{\text{full}} \times c_{\text{full}}}$. Specifically, the projection coefficient $s_i$ for each orthonormal basis vector $\vec{b}_i$ is obtained as
\begin{equation}
s_i = \vec{b}_i^{\top} \vec{x},
\end{equation}
where $\vec{b}_i^{\top} \vec{b}_j = \delta_{ij}$ holds due to orthogonality, where
\begin{equation}
\delta_{ij} =
\begin{cases}
1, & \text{if } i = j, \\
0, & \text{if } i \ne j.
\end{cases}
\end{equation}
$\delta_{ij}$ denotes the Kronecker delta function, which is 1 when $i=j$ and 0 otherwise.
\begin{figure}
    \centering
    \includegraphics[width=1.\linewidth]{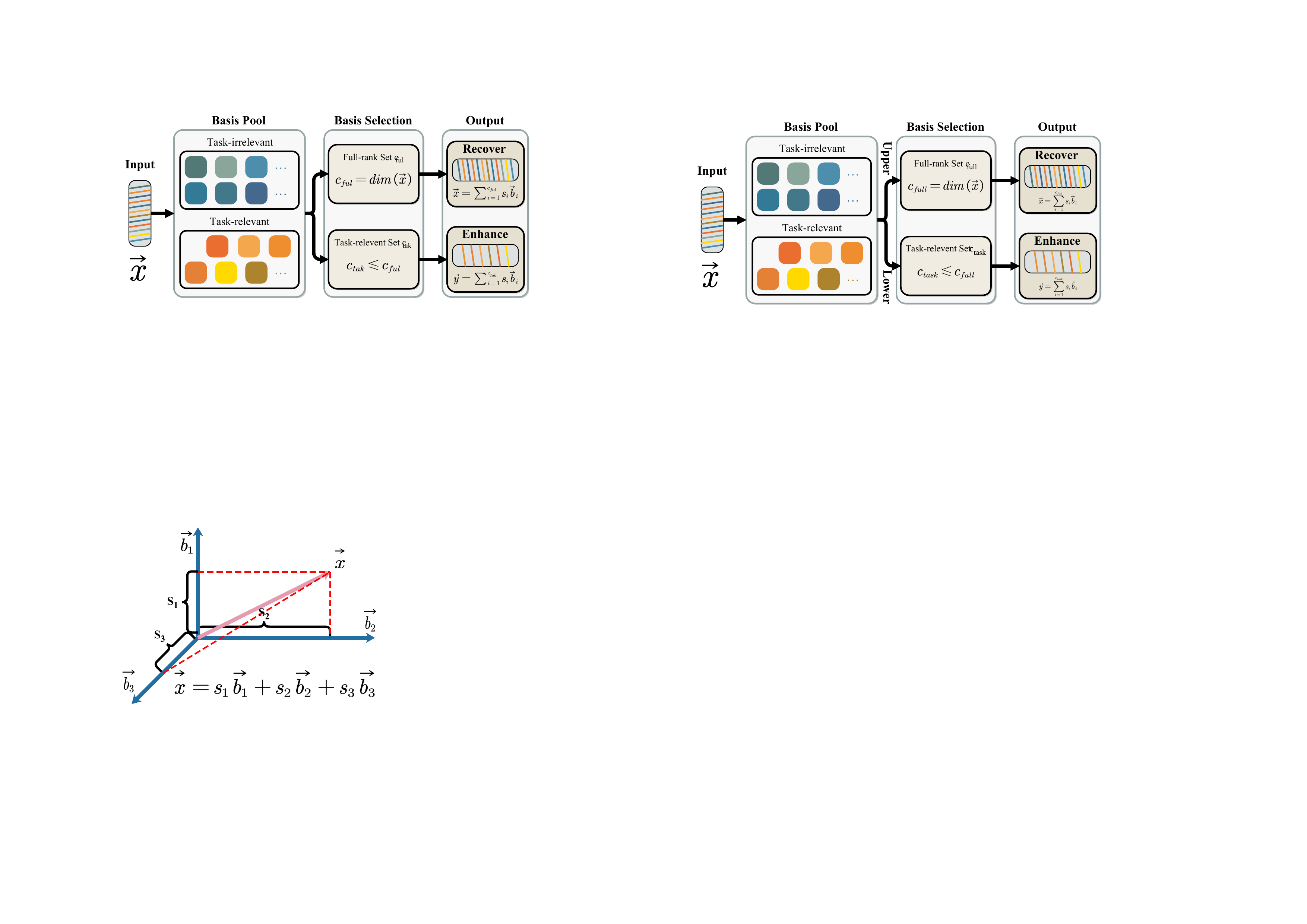}
    \caption{
Comparison between conventional full basis decomposition (upper) and our selective reconstruction (lower).
    }
    \label{basis_deomposition_framework}
\end{figure}

As illustrated in the \textbf{upper} part of Fig. \ref{basis_deomposition_framework}, a conventional full basis decomposition achieves an identity mapping:
\begin{equation}
\vec{x} = \sum_{i=1}^{c_{\text{full}}} s_i \vec{b}_i,
\end{equation}
ensuring zero information loss.

In contrast to vanilla reconstruction, we propose a selective reconstruction mechanism, as illustrated in the \textbf{lower} part of Fig. \ref{basis_deomposition_framework}. This approach identifies a task-specific subset of indices to reconstruct a refined representation $\vec{y}$ as follows:
\begin{equation}
\label{eq_2}
\vec{y} = \sum_{i=1}^{c_\text{task}} s_i \vec{b}_i, \quad \text{where} \quad s_i = \vec{b}_i^{\top} \vec{x}
\end{equation}
where $c_{\text{task}}$ is the number of the task-relevant basis features that we select. 

\subsubsection{Relaxing Constraints}
Typically, basis vectors are constrained to be orthonormal. However, strictly enforcing orthogonality, particularly in high-dimensional latent spaces, leads to considerable computational cost. In this work, we relax the orthogonality constraint. As demonstrated in Table~\ref{tab:orthogonality_ablation}, our model maintains competitive performance without orthogonality.
In contrast, we retain the unit-norm constraint
\begin{equation}
\|\vec{b}_i\|_F = 1,
\end{equation}
where $\| \cdot \|_F$ is the Frobenius norm.
Normalization constraint is critical for the following reasons:
\begin{itemize}
\item It removes the scaling ambiguity by making the basis vectors capture only structural patterns, while the coefficients are responsible for the magnitude.
\item  It prevents imbalance caused by basis vectors with excessively large magnitudes, ensuring that all basis components contribute more evenly during training.
\end{itemize}

\textbf{Thus, the orthogonality constraints are relaxed, while the normalization requirement is retained.}

 \begin{figure}
    \centering
    \includegraphics[width=.7\linewidth]{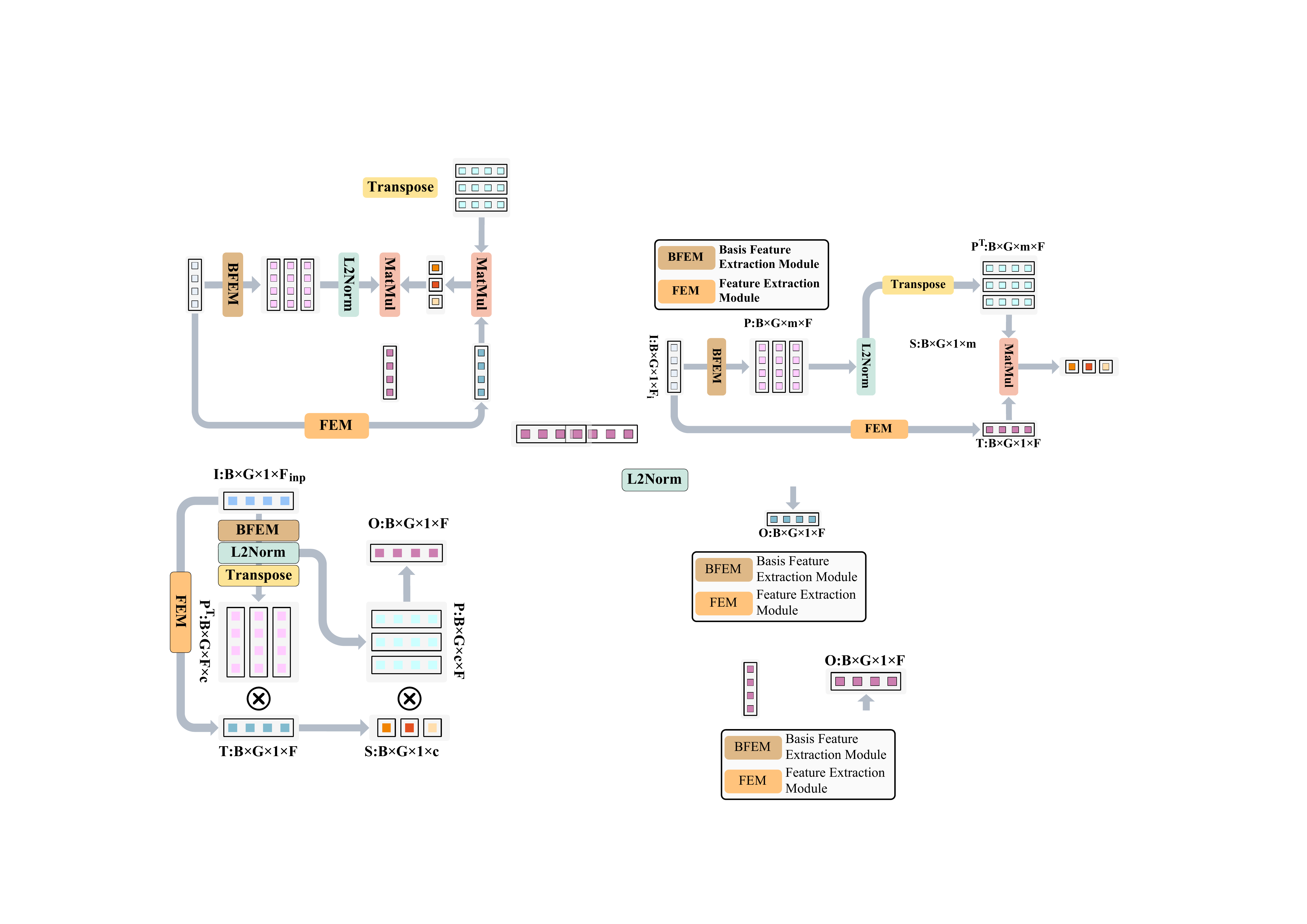}
    \caption{Overview of the BDM.}
    \label{fig:FDecM}
\end{figure}

\subsection{Basis Decomposition Module}
\label{sec:Feature Decomposition Module}
Based on the preceding discussion, we propose the Basis Decomposition Module (BDM) to process standard 4D input tensors, as illustrated in Fig. \ref{fig:FDecM}.

First, we employ a Feature Extraction Module (FEM) and a Basis Feature Extraction Module (BFEM) to obtain the original feature $T \in \mathbb{R}^{B \times G \times 1 \times F}$ and the basis features $P_i \in \mathbb{R}^{B \times G \times 1 \times F}$ from the input $I \in \mathbb{R}^{B \times G \times 1 \times F_{\text{inp}}}$: 
\begin{align}
    T &= \mathrm{FEM}(I)\,, \\
    P_1, \dots, P_{c} &= \mathrm{L2Norm}(\mathrm{BFEM}(I))\,,
\end{align}
where $B$ is the batch size, $F$ and $F_{\text{inp}}$ are the output and input feature dimensions, respectively, and $G$ represents the number of feature groups. Additionally, $c$ indicates the number of selected basis features, and $\mathrm{L2Norm}(\cdot)$ denotes L2 normalization.

Then, We concatenate the $\{P_i\}$ to obtain $P \in \mathbb{R}^{B \times G \times c \times F}$:
\begin{equation}
    P = [P_1;\dots;P_c]\,,
\end{equation}
where $[;]$ is the concatenation operator applied along the third dimension.

Subsequently, we derive the unified coefficient tensor $S \in \mathbb{R}^{B \times G \times 1 \times c}$:
\begin{equation}
    \label{deomposition}
    S = T \cdot P ^{\top}.
\end{equation}

Finally, the weight $S_i$ is assigned to each basis feature $P_i$:
\begin{equation}
     \label{reconstruction}
    O = S \cdot P= (T \cdot P^{\top})\cdot P=\sum_{i=1}^{c}(T\cdot P_i^{\top})\cdot P_i \,.
\end{equation}

To process standard 2D feature maps $\mathbf{I} \in \mathbb{R}^{B \times C \times H \times W}$, the spatial dimensions are flattened into tokens. Specifically, the spatial size $HW$ and channel dimension $C$ are mapped to $F$ and $G$, respectively, resulting in basis features of size $B \times C \times c \times (HW)$.

While the dot-product formulation of the BDM resembles the Self-Attention Mechanism, it differs fundamentally in three aspects. First, rather than applying parametric attention weighting to refine features, BDM performs linear basis reconstruction. Second, the BFEM is specifically engineered to model the unique characteristics of the applied background, ensuring the adaptability of our models for various tasks. Finally, the module is grounded in the theory of basis decomposition, focusing on signal decomposition rather than relational importance.

 \begin{figure}
    \subfloat[SGDM]{\includegraphics[width=0.5\linewidth]{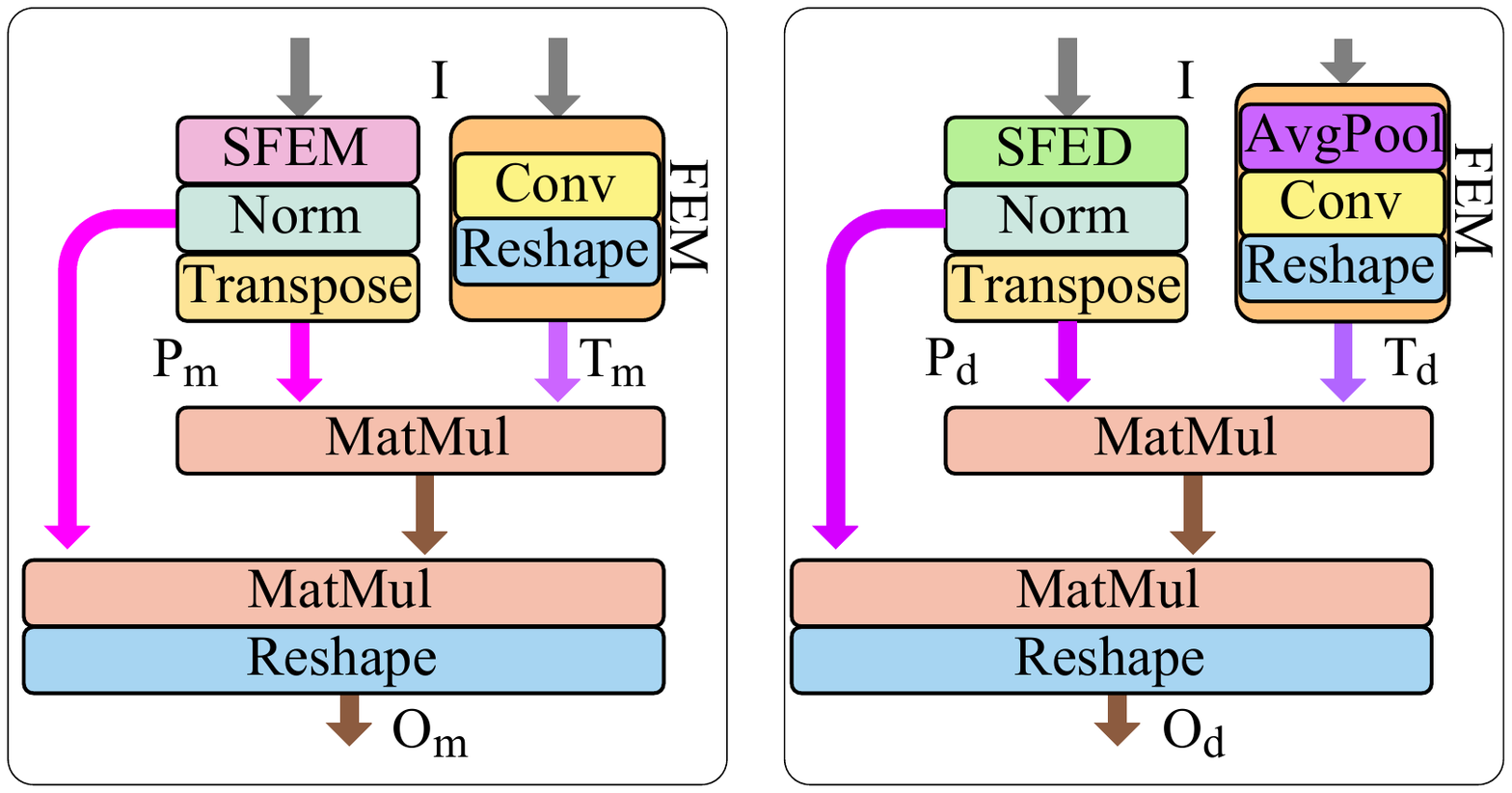}\label{SDM}}
    \subfloat[SGDDM]{\includegraphics[width=0.5\linewidth]{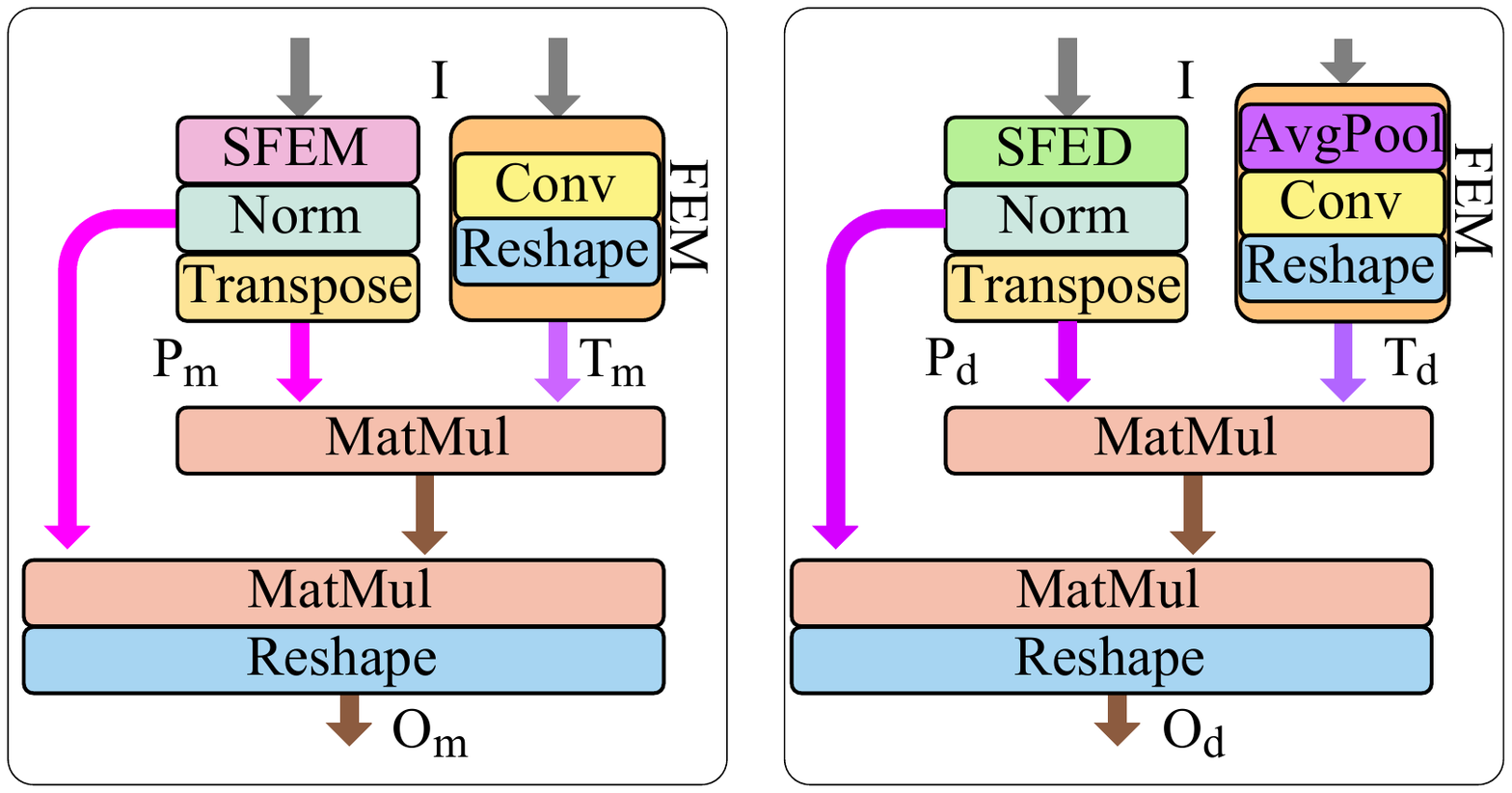}\label{SDD}}
     \caption{Detailed structures of the SGDM and SGDDM, used in Figures \ref{fig:sdiffformer} and \ref{fig:InceptionPooling}, are constructed based on the BDM. }
     \label{SDecMD}
\end{figure}

\subsection{Gradient Decomposition Module}
In IRSTD, extracting gradient-based information is fundamental, as point-like small targets are primarily characterized by localized intensity variations. 
These gradient features are computed by measuring the contrast between a central pixel and its reference pixels, which functions as a high-pass filter. 
By extracting these high-frequency components, the model can effectively distinguish small and dim target signals from the smoother and low-frequency backgrounds.
Additionally, we propose the BDM framework which could enhance the task-relevant information by decomposition-and-reconstruction structure. 
Building upon the BDM framework, we propose the GDM. 
This module utilizes spatial or temporal gradient responses as basis features to reconstruct a refined feature map. 
By ensuring the reconstructed features consist entirely of gradient responses, the GDM significantly enhances the local contrast of IRSTD targets.

Specifically, we can reformulate the reconstruction process by treating the directional gradients as a set of normalized basis vectors. First, we define the unit gradient vector $\vec{p}_i$ by normalizing the vectorized output of the derivative operator $\mathcal{D}_i$:
\begin{equation}
    \vec{p}_{i} = \frac{\operatorname{vec}(\mathcal{D}_i(\vec{x}))}{\|\operatorname{vec}(\mathcal{D}_i(\vec{x}))\|_2}, 
    \label{eq:unified_gradient_basis}
\end{equation}
Using these vectors, the enhanced signal $\vec{y}_{\text{enhance}}$ is reconstructed as a linear combination of the gradient basis features. 
The contribution of each basis, represented by the scalar weight $s_i$, is determined by the projection of the input $\vec{x}$ onto $\vec{p}_i$:
\begin{equation}
    \vec{y}_{\text{enhance}} = \sum_{i=1}^{c_{\text{grad}}} s_i \vec{p}_{i}, \quad \text{where} \quad s_i = \vec{p}_i^{\top} \vec{x},
    \label{eq:vector_reconstruction}
\end{equation}
where $c_{\text{grad}}$ is the number of gradient basis vectors determined by humans. 
As demonstrated, the enhanced features are constructed from a weighted summation of gradient basis vectors, which effectively enhance the high-frequency components associated with small targets.

As shown above, the reconstructed result $\vec{y}_{\text{enhance}}$ is the summation of the gradient vectors $\{\vec{p}_i\}_{i=1}^{c_{\text{grad}}}$, weighted by their correlation with the original input. This formulation effectively represents the signal as an orthogonal projection onto the subspace spanned by the gradient operators.

Unlike conventional gradient-based methods, the GDM offers an adaptive and interpretable architecture that adapts to target diversity:
\begin{itemize}
    \item \textbf{Dynamic Adaptability:} While standard methods often rely on fixed-weight kernels, the GDM generates input-dependent coefficients. This enables the model to adapt its response to varied target characteristics and background clutter in real-time.
    \item  \textbf{Mathematical Interpretability:} By applying a mathematically grounded decomposition theory, our approach provides a more interpretable framework.
\end{itemize}

To sum up, GDM operates by utilizing gradient responses as fundamental basis features for feature reconstruction, whose coefficients are computed.
Unlike conventional methods, GDM provides greater dynamic flexibility and mathematical interpretability by assigning adaptive weights to various directional components according to basis decomposition theory.
\begin{figure}
    \centering
    \includegraphics[width=1.\linewidth]{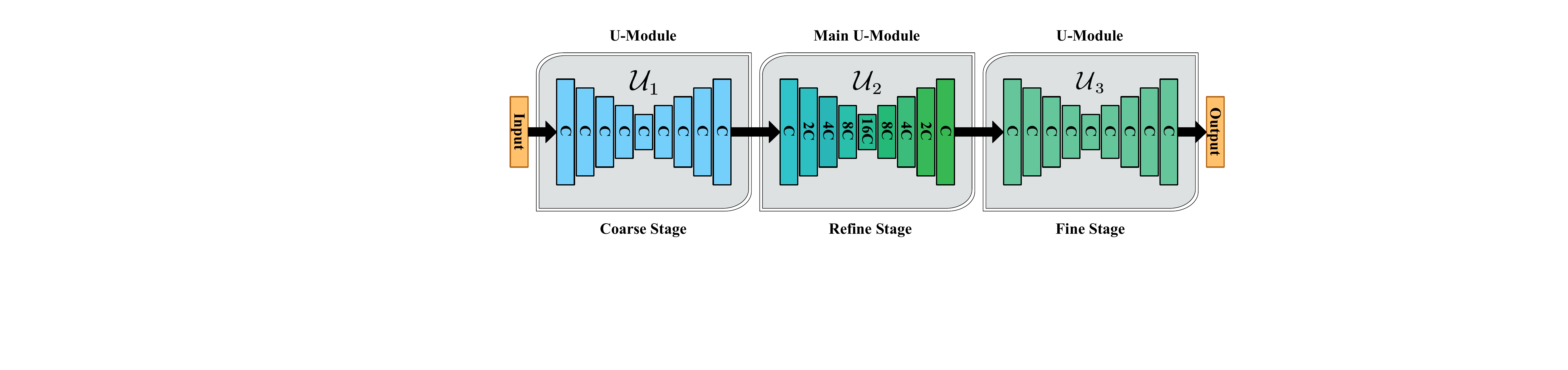}
    \caption{Our backbone is a three-stage U-Net, where each stage follows an encoder–decoder structure. $C$ denotes the number of channels in the shallowest layers. The first stage extracts coarse features, the second refines them, and the third enhances fine details.}
    \label{fig:3UNet}
\end{figure}
\begin{figure*}
    \centering
    \includegraphics[width=\linewidth]{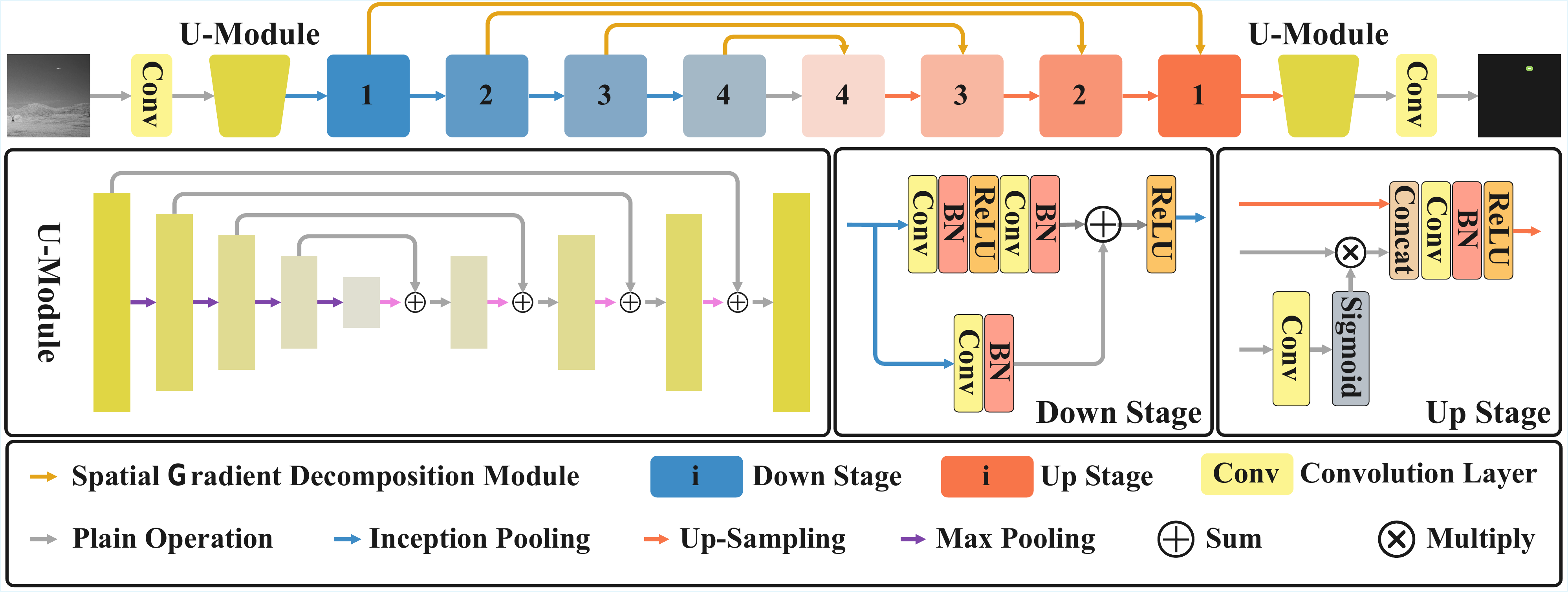}
    \caption{The overall architecture of SGBD-Net, which incorporates the SGDM and SGDDM modules.}
    \label{fig:sdiffformer}
\end{figure*}

\subsection{Spatial Gradient Basis Decomposition Network}
\label{sec:Spatial Difference Decomposition Network}
As illustrated in Figures \ref{SDM} and \ref{SDD}, we introduce two GDMs. The first, termed the Spatial Gradient Decomposition Module (SGDM), preserves the spatial resolution to enhance fine-grained target features. The second, called the Spatial Gradient Decomposition Downsampling Module (SGDDM), performs downsampling to suppress background clutter while retaining discriminative gradient information.
In addition, we integrate these modules into our backbone, as shown in Fig. \ref{fig:3UNet}, and propose the SGBD-Net, as illustrated in Fig. \ref{fig:sdiffformer}.

SGDM serves as a gradient information extractor that maintains resolution. SGDDM acts as a detail-preserving downsampling module, embedded into a multi-branch module.

\subsubsection{Three-stage U-Net}
As illustrated in Fig. \ref{fig:3UNet}, we design an architecture composed of three U-Modules. Given an input $I$, the first U-Net produces a coarse representation $f_1 = \mathcal{U}_1(I)$. The second U-Net extracts features further and works as the main feature extraction layer: $f_2 = \mathcal{U}_2(f_1)$. The third U-Module works for final result enhancement: $O = \mathcal{U}_3(f_2)$. The overall mapping is
\begin{equation}
O = \mathcal{U}_3\big(\mathcal{U}_2(\mathcal{U}_1(I))\big).    
\end{equation}
The motivation for this specific configuration stems from a fundamental trade-off between computational efficiency and representational capacity. Specifically, the integration of the GDM introduces a computational burden that increases with the number of channels. To maintain a competitive FLOP count and ensure real-time viability, we compress the channel dimensions within the backbone to maintain efficiency.
To address the above challenges, we transform from a "\textbf{wide and shallow}" architecture to a "\textbf{slim and deep}" iterative strategy. 
By adopting a three-stage backbone, we enable the model to perform continuous feature refinement and stronger multi-scale aggregation. 
This iterative approach effectively compensates for the reduced channel width, allowing the network to achieve superior feature representation while remaining more computationally efficient than a single-stage and wider U-Net.
To sum up, the three-stage U-Net and GDM play complementary roles. The three-stage U-Net extracts rich semantics to build the global context. Within this structure, the GDM could filter the useless information and enhance task-relevant features. Additionally, he three-stage U-Net uses a 'slim and deep' design to manage the computational costs caused by GDM.

\begin{figure}
    \centering
    \includegraphics[width=1.\linewidth]{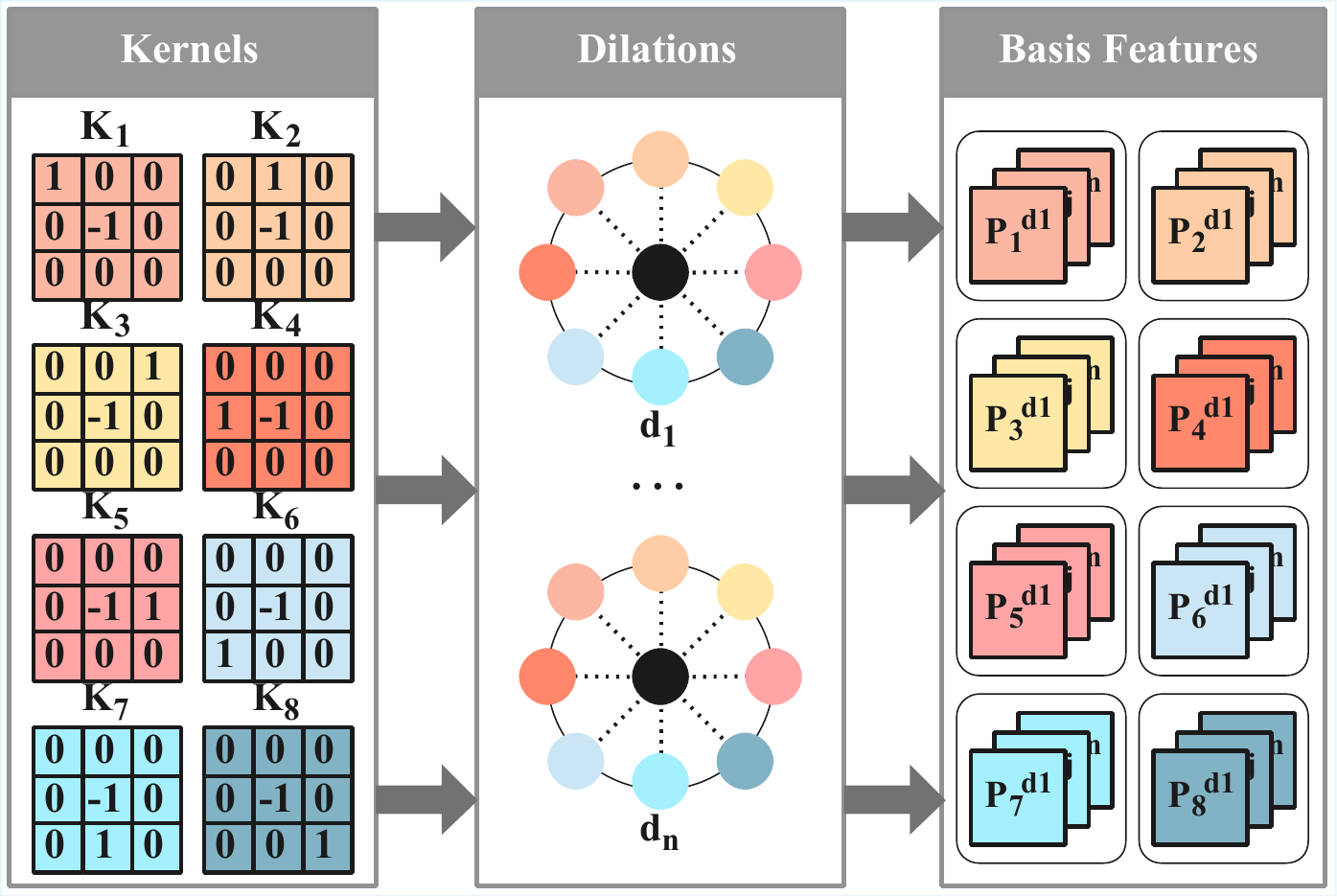}
    \caption{Architecture of the SFEM, which employs eight gradient kernels with $n$ dilation rates to extract multi-scale basis features.}
    \label{fig:DFEM}
\end{figure}

\subsubsection{Spatial Gradient Decomposition Module}
\label{SDDM_Section}
As depicted in Fig. \ref{SDM}, we employ a $1\times 1$ convolution layer followed by a reshape operation ($\mathrm{Reshape}$) as FEM to extract the original feature $T_m \in \mathbb{R}^{B\times C\times 1\times (HW)}$:
\begin{align}
    T_m = \mathrm{Reshape}(\mathrm{Conv}_{1\times 1}(I))\,,
    \label{alg:computing_T}
\end{align}
where H, W, and C denote the height, width, and channel dimension.
The Spatial Feature Extraction Module (SFEM) utilizes gradient kernels to extract high-frequency basis features ${P^{d_j}_i} \in \mathbb{R}^{B\times C\times 1\times (HW)}$ (where i and j index the kernel type and dilation rate, respectively). These features are then used to form the tensor $P_m \in \mathbb{R}^{B\times C\times (8n)\times (HW)}$:
\begin{align}
    P_m &= \mathrm{L2Norm}(SFEM(I))\,,\\
    SFEM(I) &= [P^{d_1}_1;\ldots;P^{d_n}_1;\ldots;P^{d_n}_8]\,,\\
    P^{d_j}_i &= \text{Reshape}(\mathrm{Conv}_{K_i}^{d_j}(I))\,,
\end{align}
where $\mathrm{Conv}_{K_i}^{d_j}$ is the convolution with kernels $K_i$ ($i \in \{1, \dots, 8\}$), as shown in Fig. \ref{fig:DFEM}, and dilation rates $d_j$ ($j \in \{1, \dots, n\}$) to extract the high-frequency features as our basis features.

We utilize $P_m$ and $T_m$ to obtain $O_m\in \mathbb{R}^{B\times C\times H\times W}$ according to Equation \ref{reconstruction}:
\begin{align}
    O_m = \mathrm{Reshape}((T_m\cdot P_m^{\top})\cdot P_m\,)\,,
\end{align}
 where $O_m$ is a combination of gradient features. In the process, the background information is suppressed by overlooking the low-frequency basis features.

\begin{figure}
    \centering
    \includegraphics[width=1.\linewidth]{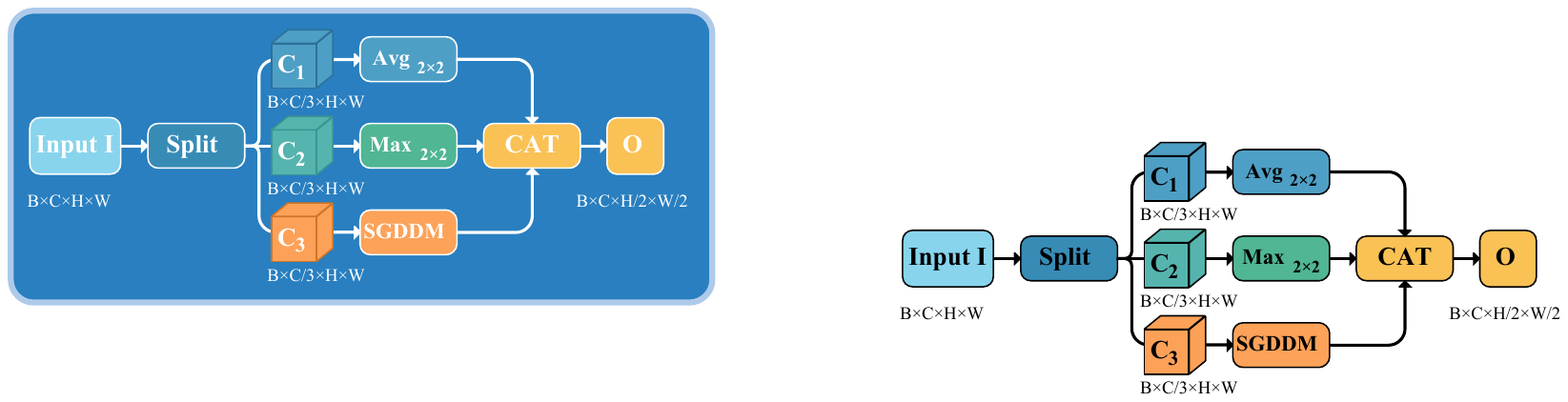}
    \caption{The architecture of Inception Pooling.}
    \label{fig:InceptionPooling}
\end{figure}

\subsubsection{Inception Pooling}
While SGDDM could keep detailed information, the downsampling process itself requires broader contextual information beyond these details.
Therefore, we utilize multiple branches to maintain various information.

To maintain feature diversity at various levels, Inception Pooling splits the feature channels into groups and applies distinct operations to each group, as shown in Fig. \ref{fig:InceptionPooling}.

First, we split the input $I$  along the channel dimension into three equal-sized tensors $C_1, C_2, C_3 \in \mathbb{R}^{B \times (C/3) \times H \times W}$.
\begin{align}
    C_1, C_2, C_3 = \mathrm{Split}(I)\,,
\end{align}

After that, we use MaxPooling $\mathrm{Max}_{\text{2×2}}$, AvgPooling $\mathrm{Avg}_{\text{2×2}}$, and SGDDM to downsample the input and concatenate the results along the channel dimension using a concatenation operation (CAT). 
\begin{align}
    O = \text{CAT}(\mathrm{Max}_{\text{2×2}}(C_1),\mathrm{Avg}_{\text{2×2}}(C_2), \text{SGDDM}(C_3))\,,
\end{align}

\subsubsection{Spatial Gradient Decomposition Downsampling Module.}
As illustrated in Fig. \ref{SDD}, the SGDDM shares a similar architecture with SGDM, differing in the original feature $T_d \in \mathbb{R}^{B\times C\times 1\times (HW/4)}$ and the basis features $P_d \in \mathbb{R}^{B\times C\times 4\times (HW/4)}$, which is computed as follows:
\begin{align}
    T_d &= \mathrm{Reshape}(\mathrm{Conv}_{\mathrm{1\times 1}}(\mathrm{Avg}_{\mathrm{2\times 2}}(I)))\,,\\
     P_d &= \mathrm{L2Norm}(\text{SFED}(I))\,,
\end{align}
where SFED denotes the Spatial Feature Extraction Downsampling, formulated as follows:
\begin{align}
    \text{SFED}(I) &= [P_{max}; P_{D_1}; P_{D_2}; P_{D_3}]\,,\\
     P_{max} &= \mathrm{Reshape}(\mathrm{Max}_{\mathrm{2\times 2}}(I))\,,\\
      P_{D_k} &= \mathrm{Reshape}(\mathrm{Conv}^1_{D_k}(I))\,,
\end{align}
where $\mathrm{Conv}^{1}_{D_k}$ denotes a convolution with stride 2, dilation ratio of 1, and the kernel $D_k$, which is defined as: $D_1=[ 1, 1; -1, -1], D_2=[ 1, -1; 1, -1], D_3=[ 1, -1; -1, 1]$.

At last, we obtain the output of $O_d\in \mathbb{R}^{B\times C\times H/2\times W/2}$:
\begin{align}
\label{alg:P_1}
    O_d = \mathrm{Reshape}((T_d\cdot P_d^{\top})\cdot P_d\,)\,.
\end{align}
where $O_d$ is a linear sum of high-frequency features and MaxPooling results from I.

\begin{figure}
\centering
\includegraphics[width=\linewidth]{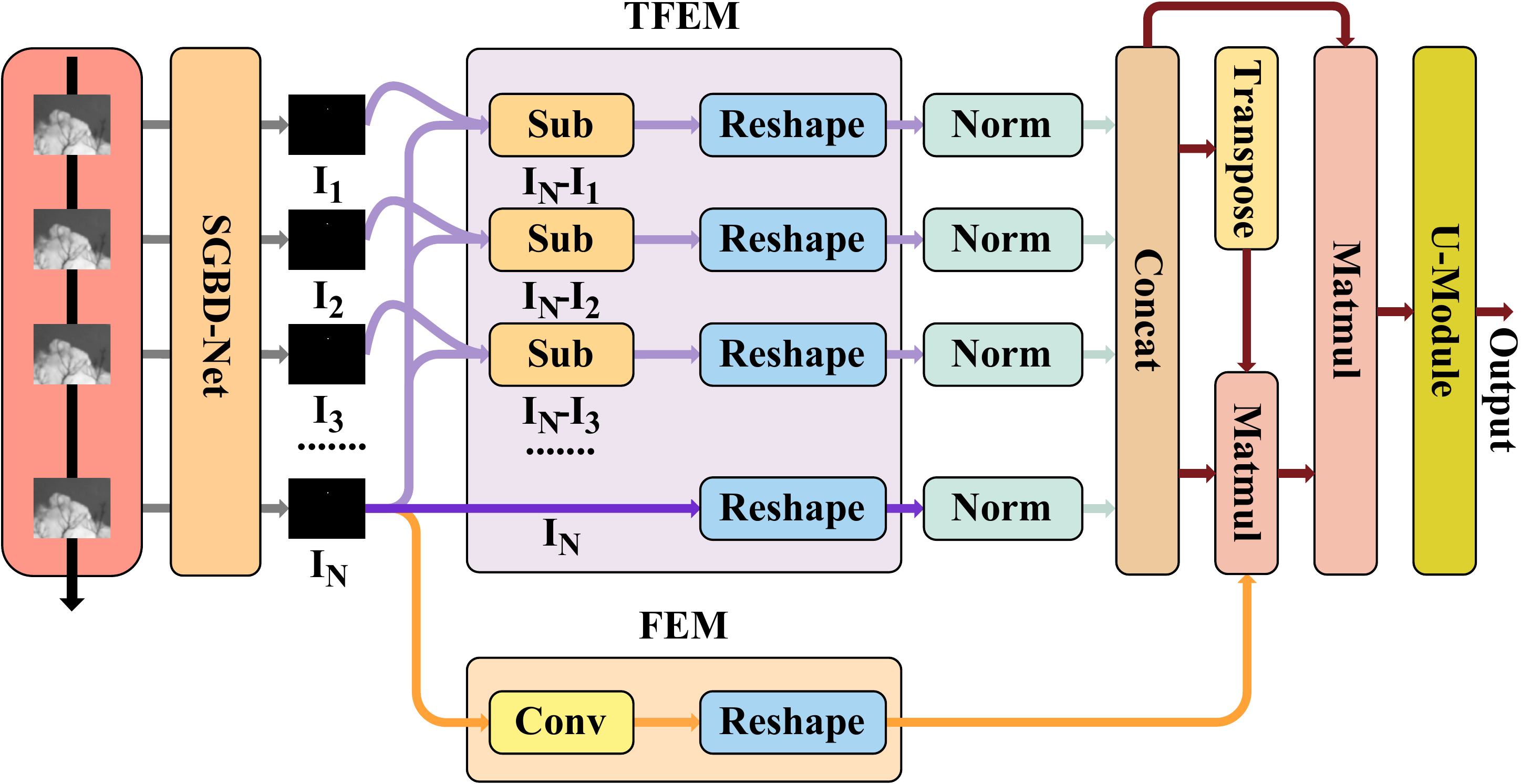}
\caption{The structure of STGBD-Net. The TFEM extracts temporal gradient basis features, while the FEM fuses reference frames with the key frame to obtain temporally enhanced features.}
\label{fig:STDecNet}
\end{figure}

\subsection{Spatio-temporal Gradient Basis Decomposition Network}
We introduce the Temporal Gradient Decomposition Module (TGDM) to suppress the strong background, and propose STGBD-Net, as shown in Fig. \ref{fig:STDecNet}. 

First, we extract the input features $I_k \in \mathbb{R}^{B\times C\times H \times W}$ (k = 1,...,$N$) for each of the $N$ consecutive frames using SGBD-Net, and then utilize the FEM to obtain original features $T_t\in \mathbb{R}^{B\times C\times 1 \times (HW)}$:
\begin{equation}
     T_t=\mathrm{Reshape}(\mathrm{Conv}_{1\times 1}(I_N))\,.
 \end{equation}
 
Next, we employ the Temporal Feature Extraction Module (TFEM) and the normalization layer to obtain the basis features $P_t \in \mathbb{R}^{B\times C\times N \times (HW)}$:
\begin{align}
    P_t &= \mathrm{L2Norm}(\mathrm{TFEM}(I_1,\dots,I_N)) \,,
     \\
   \mathrm{TFEM}(I) &= [P_{D_{1}};\dots;P_{D_{N-1}};P_C]\,,
    \\
   P_{D_{i}} &= \mathrm{Reshape}(I_N - I_{i})\,,
   \\
   P_C &= \mathrm{Reshape}(I_N),
\end{align}
where $P_{D_{i}}$ denotes the gradient between the current frame $I_N$ and the reference frame $I_i$. $P_C$ is the state of the current frame. 
They could ensure the result is the fusion of current and temporal information.

Later, we approximate $T_t$ with a combination of $P_t$ and obtain the result $O_t\in \mathbb{R}^{B\times C\times H\times W}$:
\begin{equation}
    O_t =\mathrm{Reshape}((T_t \cdot P_t^\top) \cdot P_t)\,.
\end{equation}

Finally, we use a U-Module in Fig. \ref{fig:sdiffformer} to further extract information and a convolution layer to adjust the output channel dimension.

\subsection{Loss Function}
We employ the soft Intersection over Union (IoU) loss for training our model, which is defined as:
\begin{equation}
\mathcal{L} = 1 - \frac{\sum_{i,j} p_{i,j} \cdot g_{i,j}}{\sum_{i,j} p_{i,j} + \sum_{i,j} g_{i,j} - \sum_{i,j} p_{i,j} \cdot g_{i,j}}
\label{eq:softiou}
\end{equation}
where $g_{i,j} \in [0,1]$ and $p_{i,j} \in [0,1]$ represent the ground truth and predicted probability after Sigmoid at $(i,j)$. 

\section{Experiments}
We present the experiment details, validate modules via ablation studies, and demonstrate the performance of BDM's adaptations through comparisons on both SIRSTD and MIRSTD benchmarks.

\subsection{Experimental Settings}

\subsubsection{Datasets}
To evaluate our approach, we utilize several widely recognized datasets for SIRSTD and MIRSTD.
\begin{itemize}
    \item \textbf{SIRSTD:} We evaluate our method on two single-frame datasets: NUDT-SIRST \cite{RenLiHanShu2021DNANet} and IRSTD-1K \cite{Zhang2022ISNet}.
    \item \textbf{MIRSTD:} We use NUDT-MIRSDT \cite{Li2025dtum} and 58 selected IRDST \cite{SunBaiYangBai2023Receptive-Field} sequences following \cite{huang2024lmaformer}.
\end{itemize}
The dataset splits are as follows: NUDT-MIRSDT and MIRSTD use an 80:20 training-to-test ratio; NUDT-SIRSTD follows a 50:50 split; and IRDST uses a 70:30 split. 

\subsubsection{Implementation Details}
All experiments are implemented on Ubuntu 20.04 LTS. To ensure the robustness and generalizability of our results, we evaluate the proposed algorithm across both SIRSTD and MIRSTD tasks.
\begin{itemize}
    \item \textbf{SIRSTD:} Models were trained on a single NVIDIA GeForce RTX 4080 GPU for 400 epochs. We used the Adam optimizer with an initial learning rate of $5\times10^{-4}$. The learning rate was decayed to $5\times10^{-5}$ at epoch 200 and further reduced to $5\times10^{-6}$ in the final stage. The batch size was set to 4.
    \item  \textbf{MIRSTD:} Our architecture was trained on two 24GB NVIDIA GeForce RTX 3090 GPUs for 20 epochs. We employed the Adam optimizer with an initial learning rate of $1\times10^{-3}$ and a batch size of 4. The temporal window size $N$ for STGBD-Net was set to 5, which is commonly applied in many papers \cite{ying2025rfr, Li2025dtum, Deng2026DQAligner}.
\end{itemize}
Furthermore, we assess the computational efficiency of our proposed methods using a standardized input resolution of $256\times256$ pixels and GeForce RTX 4080 GPU for both SIRSTD and MIRSTD.

\subsubsection{Evaluation Metrics}
We assess model performance using both accuracy-based and efficiency-based metrics:
\begin{itemize}
    \item \textbf{Accuracy Metrics:} Probability of Detection (Pd), False Alarm Rate (Fa), Mean Intersection-over-Union (mIoU), and Area Under the Curve (AUC).
    \item \textbf{Efficiency Metrics:} Frame Per Second (FPS), Giga Floating-point Operations Per Second (GFLOPS), and Parameters (Params).
\end{itemize}

\begin{table}
\centering
\caption{Effectiveness of orthogonality and normalization.}
\begin{adjustbox}{max width=\linewidth}
\begin{threeparttable}
\setlength{\tabcolsep}{3pt}
\renewcommand{\arraystretch}{1.15}


\begin{tabular}{c|c||cc|cc}
\toprule[1.5pt]
\multirow{2}{*}{\bfseries Normalization} \
&\multirow{2}{*}{\bfseries Orthogonality} 
&\multicolumn{2}{c|}{\bfseries NUDT\mbox{-}SIRST} 
 &\bfseries FPS $\uparrow$
\\
& &mIoU(\%)$\uparrow$ & AUC(\%)$\uparrow$
 &frames/s
\\
\midrule
\midrule
\checkmark &\checkmark & \underline{94.14} & \underline{98.37} &16.94 \\
\rowcolor{headbg}
- &-  &92.70  &97.85 &\textbf{224.32} \\
\checkmark &-  & \textbf{95.13 }& \textbf{98.96} & \underline{217.06} \\
\toprule[1.5pt]
\end{tabular}
\end{threeparttable}
\end{adjustbox}
\label{tab:orthogonality_ablation}
\end{table}

\begin{table}
\centering
\caption{Effectiveness of SGDM's dilation configurations on NUDT-SIRST. We bold the best results and underline the second best.}
\begin{adjustbox}{max width=\linewidth}
\begin{threeparttable}
\setlength{\tabcolsep}{3pt}
\renewcommand{\arraystretch}{1.15}


\begin{tabular}{c||cc|cc}
\toprule[1.5pt]
\multirow{2}{*}{\bfseries Dilation Ratios} 
&\multicolumn{2}{c|}{\bfseries NUDT\mbox{-}SIRST} 
&\bfseries Params$\downarrow$ &\bfseries FPS $\uparrow$
\\
&mIoU(\%)$\uparrow$ & AUC(\%)$\uparrow$
&(M) &frames/s
\\
\midrule
\midrule
{[1]} &94.33  &97.34  &\textbf{0.23}  &\textbf{230.04} \\
\rowcolor{headbg}
{[1,2]} &\underline{94.65} &\underline{98.74} &\underline{0.27} &\underline{218.27} \\
{[1,2,3]}  & \textbf{95.13} & \textbf{98.96} & 0.27 & 217.06 \\
\rowcolor{headbg}
{[1,2,3,4]} &94.23  &97.58 &0.37 &201.00 \\
\toprule[1.5pt]
\end{tabular}
\end{threeparttable}
\end{adjustbox}
\label{tab:dilation_ablation}
\end{table}
\begin{table}
\centering
\caption{Effectiveness of SGBD-Net's channel configurations on NUDT-SIRST. We bold the best results and underline the second best.}
\begin{adjustbox}{max width=\linewidth}
\begin{threeparttable}
\setlength{\tabcolsep}{3pt}
\renewcommand{\arraystretch}{1.15}


\begin{tabular}{c||cc|cc}
\toprule[1.5pt]
\multirow{2}{*}{\bfseries Channels} 
&\multicolumn{2}{c|}{\bfseries NUDT\mbox{-}SIRST} 
&\bfseries Params$\downarrow$ &\bfseries FPS $\uparrow$
\\
&mIoU(\%)$\uparrow$ & AUC(\%)$\uparrow$
&(M) &frames/s
\\
\midrule
\midrule
{[4,8,16,32]} & 94.14 & 97.53 & \textbf{0.07} & \underline{217.41} \\
\rowcolor{headbg}
{[8,16,32,64]} & 95.13 & \underline{98.96} & \underline{0.27} & \textbf{217.06} \\
{[16,32,64,128]} & \underline{95.56} & \textbf{99.17} & 1.03 & 211.20 \\
\rowcolor{headbg}
{[32,64,128,256]} & \textbf{96.00} & 98.29 & 4.11 & 171.85 \\
\toprule[1.5pt]
\end{tabular}
\end{threeparttable}
\end{adjustbox}
\label{tab:channel_ablation}
\end{table}

\begin{table}
\centering
\caption{Comparative analysis of our decomposition‑based module with other modules on the SIRSTD dataset.}
\begin{adjustbox}{max width=\linewidth}
\begin{threeparttable}
\setlength{\tabcolsep}{3pt}
\renewcommand{\arraystretch}{1.15}


\begin{tabular}{c||cc|cc}
\toprule[1.5pt]

\multirow{2}{*}{\bfseries Methods} 
&\multicolumn{2}{c|}{\bfseries NUDT\mbox{-}SIRST} 
&\multicolumn{2}{c}{\bfseries IRSTD\mbox{-}1K} 
\\
&mIoU(\%)$\uparrow$ & $\mathrm{AUC}(\%)\uparrow$
&mIoU(\%)$\uparrow$ & $\mathrm{AUC}(\%)\uparrow$
\\
\midrule
\midrule
Haar &\underline{92.46} &\underline{98.45} &62.23 &87.46 \\
\rowcolor{headbg}Fourier &90.60 & 96.84 &\underline{62.64} & \textbf{89.71}\\
SGDM &\textbf{95.13} &\textbf{98.96} &\textbf{69.40} &\underline{89.68}\\
\toprule[1.5pt]
\end{tabular}

\end{threeparttable}
\end{adjustbox}
\label{Quantitative Module Comparison}
\end{table}
\begin{table}
\centering
\caption{Comparative analysis of our method against self‑attention‑based dynamic convolutions.}
\begin{adjustbox}{max width=\linewidth}
\begin{threeparttable}
\setlength{\tabcolsep}{3pt}
\renewcommand{\arraystretch}{1.15}


\begin{tabular}{c||c|cccc|c}
\toprule[1.5pt]

Methods
&Venue 
&mIoU(\%)$\uparrow$ 
& $\mathrm{P_d}(\%)\uparrow$ 
& $\mathrm{F_a}(10\textsuperscript{-6})\downarrow$ 
& $\mathrm{AUC}(\%)\uparrow$
&FPS(rames/s))$\uparrow$
\\
\midrule
\midrule
ODConv &ICLR\textsuperscript{22}&93.56 &97.21 &6.89 &98.21 &\underline{168.83} \\
\rowcolor{headbg}DCNv4 &CVPR\textsuperscript{24}&\underline{94.20} &\underline{97.35}  &5.49 &\underline{98.75} &120.18 \\
FDConv &CVPR\textsuperscript{25}&93.96 & 97.18 &\underline{5.40} & 98.42 &128.37\\
\rowcolor{headbg}SGDM &-&\textbf{95.13} &\textbf{97.88} &\textbf{2.50} &\textbf{98.96} & \textbf{217.06}  \\
\toprule[1.5pt]
\end{tabular}

\end{threeparttable}
\end{adjustbox}
\label{Quantitative dynamic convolution Comparison}
\end{table}
\begin{table*}
\centering
\caption{Ablation studies of each module in SGBD-Net using SIRSTD datasets. We bold the best results and underline the second best.}
\begin{adjustbox}{max width=\linewidth}
\begin{threeparttable}
\setlength{\tabcolsep}{3pt}
\renewcommand{\arraystretch}{1.15}


\begin{tabular}{ccc||cccc|cccc|cc}
\toprule[1.5pt]
\multirow{2}{*}{\bfseries SGDM}
&\multirow{2}{*}{\bfseries InceptionPool}
&\multirow{2}{*}{\bfseries SGDDM}
&\multicolumn{4}{c|}{\bfseries NUDT-SIRST}
&\multicolumn{4}{c|}{\bfseries IRSTD-1K}
&\bfseries FPS$\uparrow$\\
&&\MetricHead \MetricHead &(frames/s)\\
\midrule
\midrule
$-$ & $-$ &$-$ &92.20(-)&97.67(-) &10.75(-) &97.94(-) & 66.96(-) & 88.89(-)& 15.32(-)& 86.62(-) &\textbf{374.33}(-)\\
\rowcolor{headbg}
$\checkmark$ & $-$ &$-$ &\textbf{95.21}(+3.01) &\underline{98.09}(+0.42) &\textbf{0.94}(+9.81) &98.54(+0.60) &\underline{68.01}(+1.05) &90.90(+2.01) &14.27(+1.05) &\textbf{90.97}(+4.35) &268.70(-105.63) \\
$\checkmark$ & $\checkmark$ &$-$ &94.99(+2.79) &\textbf{98.30}(+0.63)&3.38(+7.37) &98.60(+0.66)&67.54(+0.58) &88.92(+0.03) &\textbf{12.32}(+3.00) & 89.92(+3.30) &258.90(-115.43) \\
\rowcolor{headbg}
 $-$ &$\checkmark$&$-$ &93.83(+1.63) &97.28(-0.39) &2.99(+7.76)&98.22(+0.28)  &66.10(-0.86) &87.20(-1.69) &16.32(-1.00) & 88.05(+1.43) &\underline{336.82}(-37.51)\\
 $-$ &$\checkmark$&$\checkmark$ &94.61(+2.41) &97.77(+0.10) & \underline{1.86}(+8.89) & \underline{98.61}(+0.67) &67.46(+0.50) &\underline{91.24}(+2.35) &17.18(-1.86) & \underline{90.77}(+4.15)&273.20(-101.13) \\
 \rowcolor{headbg}
 $\checkmark$ &$\checkmark$&$\checkmark$  &\underline{95.13}(+2.93) &97.88(+0.21) &2.50(+8.25) &\textbf{98.96}(+1.02) &\textbf{69.40}(+2.44) &\textbf{91.26}(+2.37) &\underline{13.63}(+1.69) &89.68(+3.06) &217.06(-157.27)\\
\toprule[1.5pt]
\end{tabular}
\end{threeparttable}
\end{adjustbox}
\label{table_ablation_sdecnet}
\end{table*}
\begin{figure*}
    \centering
    \subfloat[NUDT-SIRST]{\includegraphics[width=0.23\textwidth]{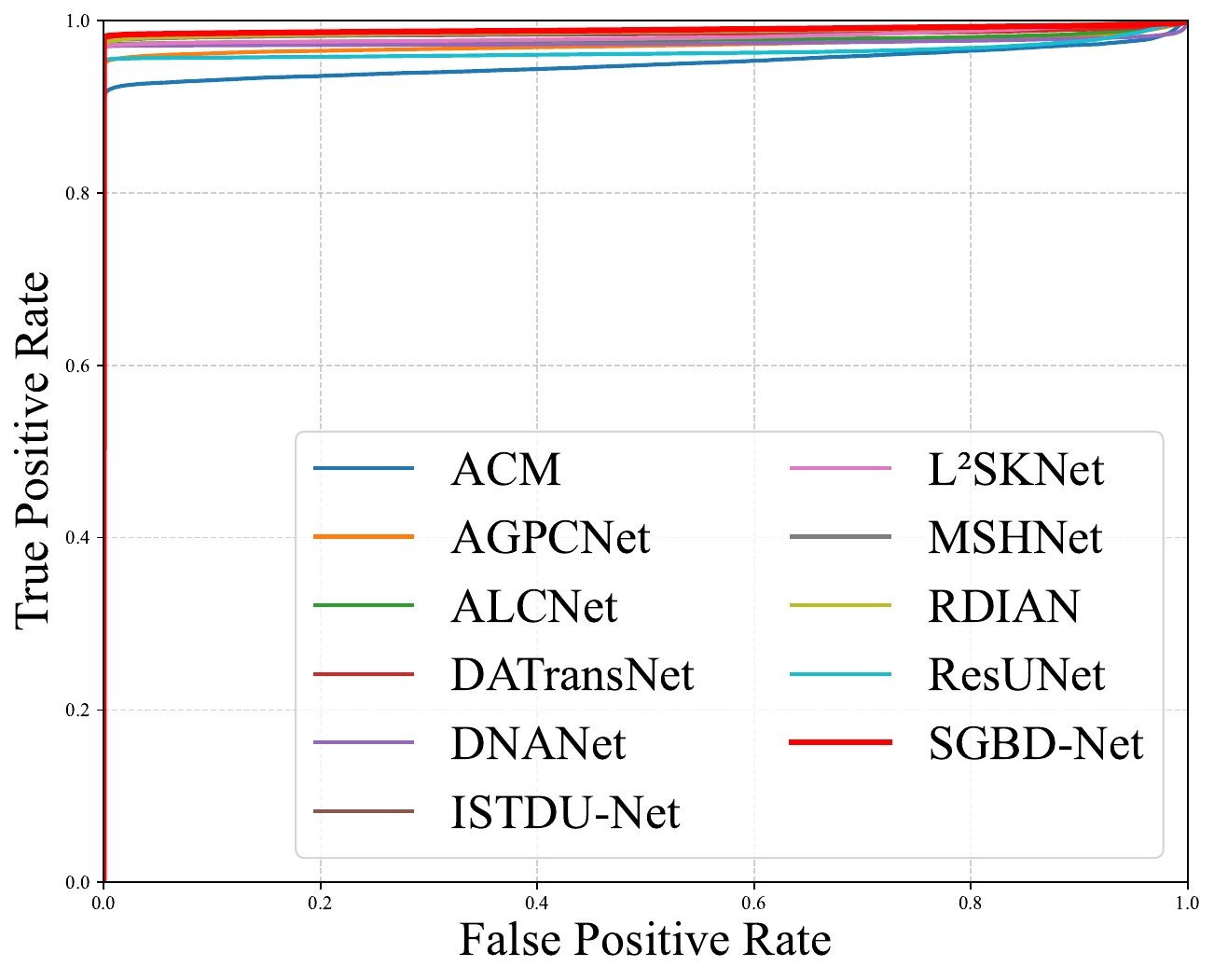}}%
    \hfill
    \subfloat[IRSTD-1K]{\includegraphics[width=0.23\textwidth]{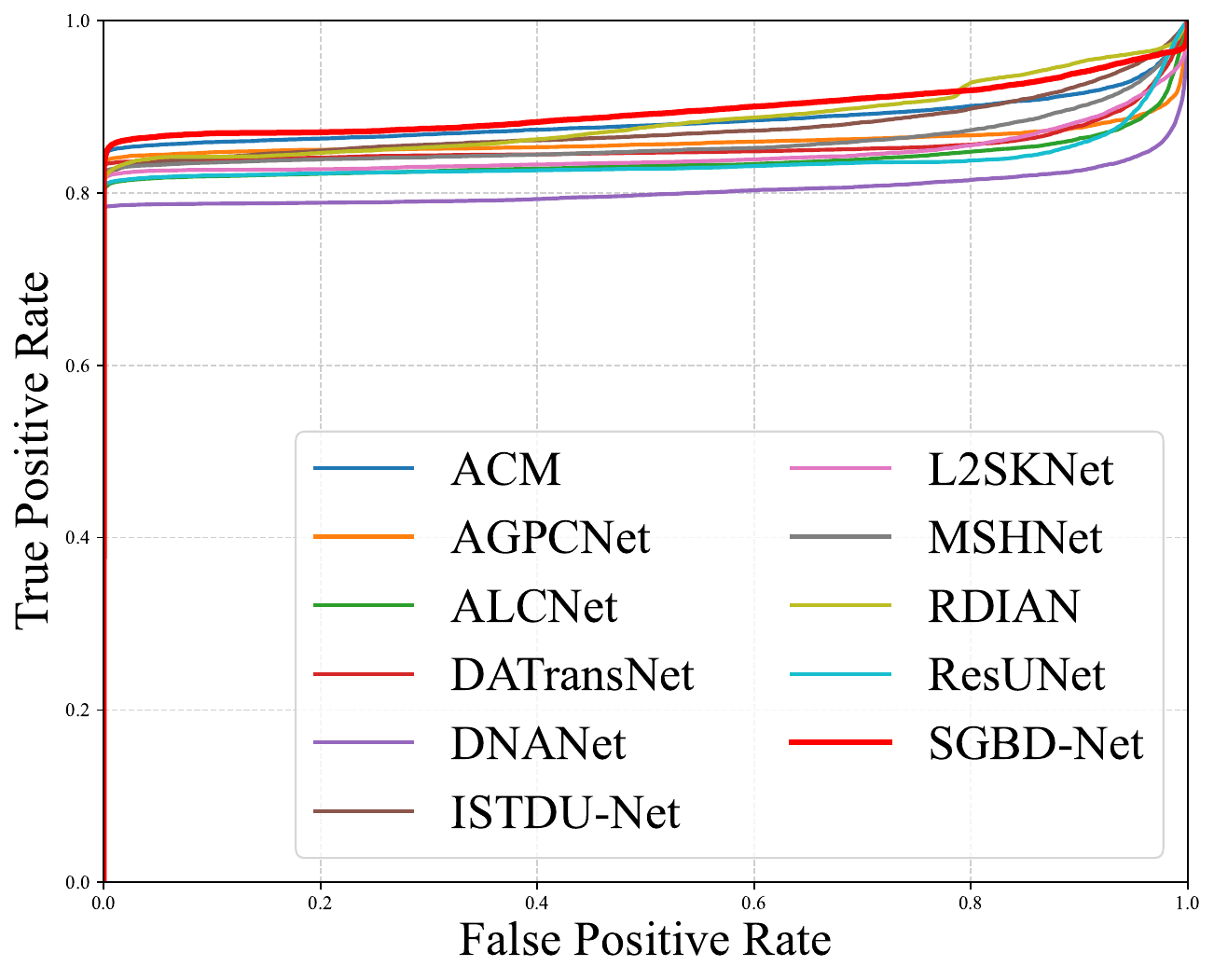}}%
    \hfill
    \subfloat[Zoomed NUDT-SIRST]{\includegraphics[width=0.23\textwidth]{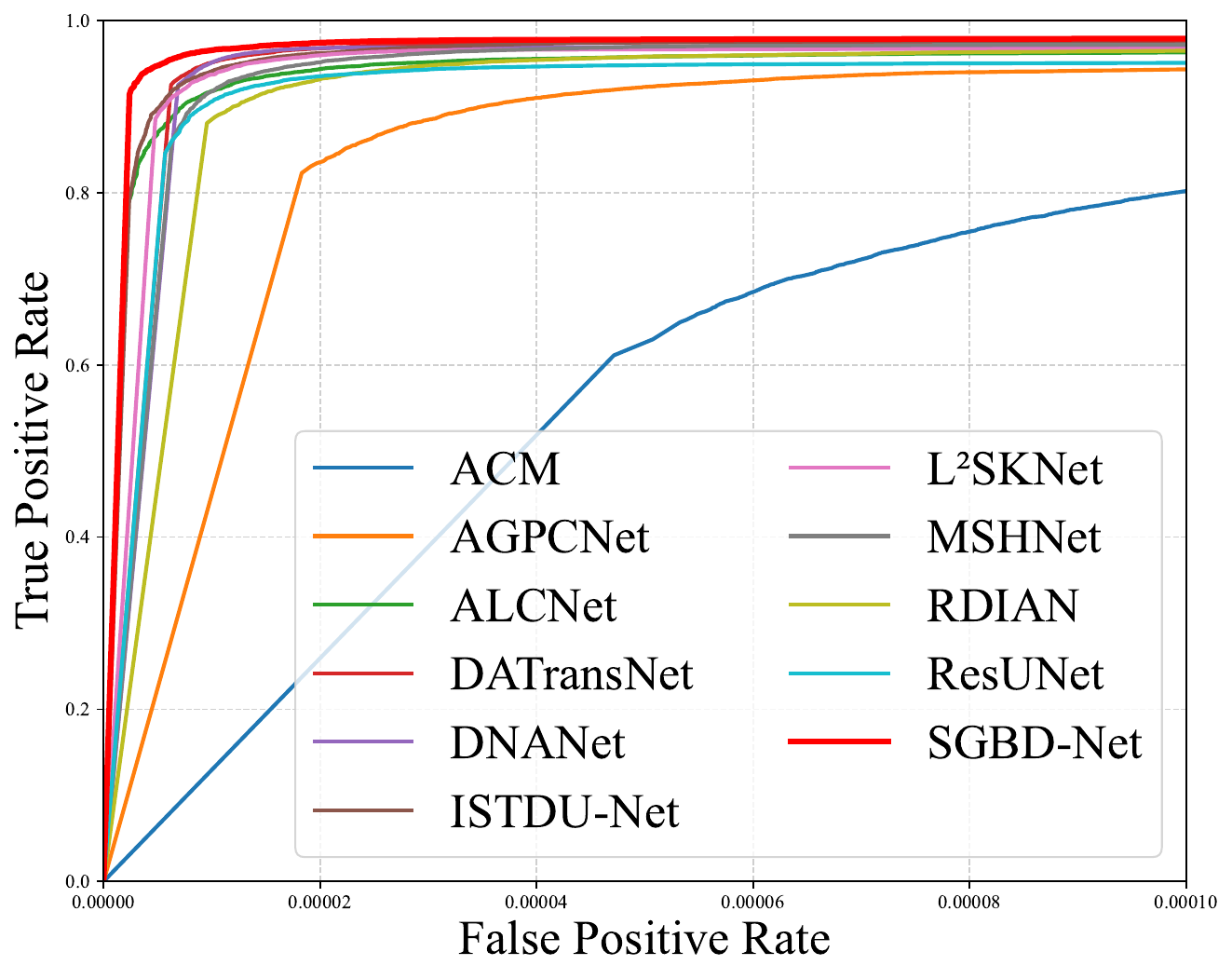}}%
    \hfill
    \subfloat[Zoomed IRSTD-1K]{\includegraphics[width=0.23\textwidth]{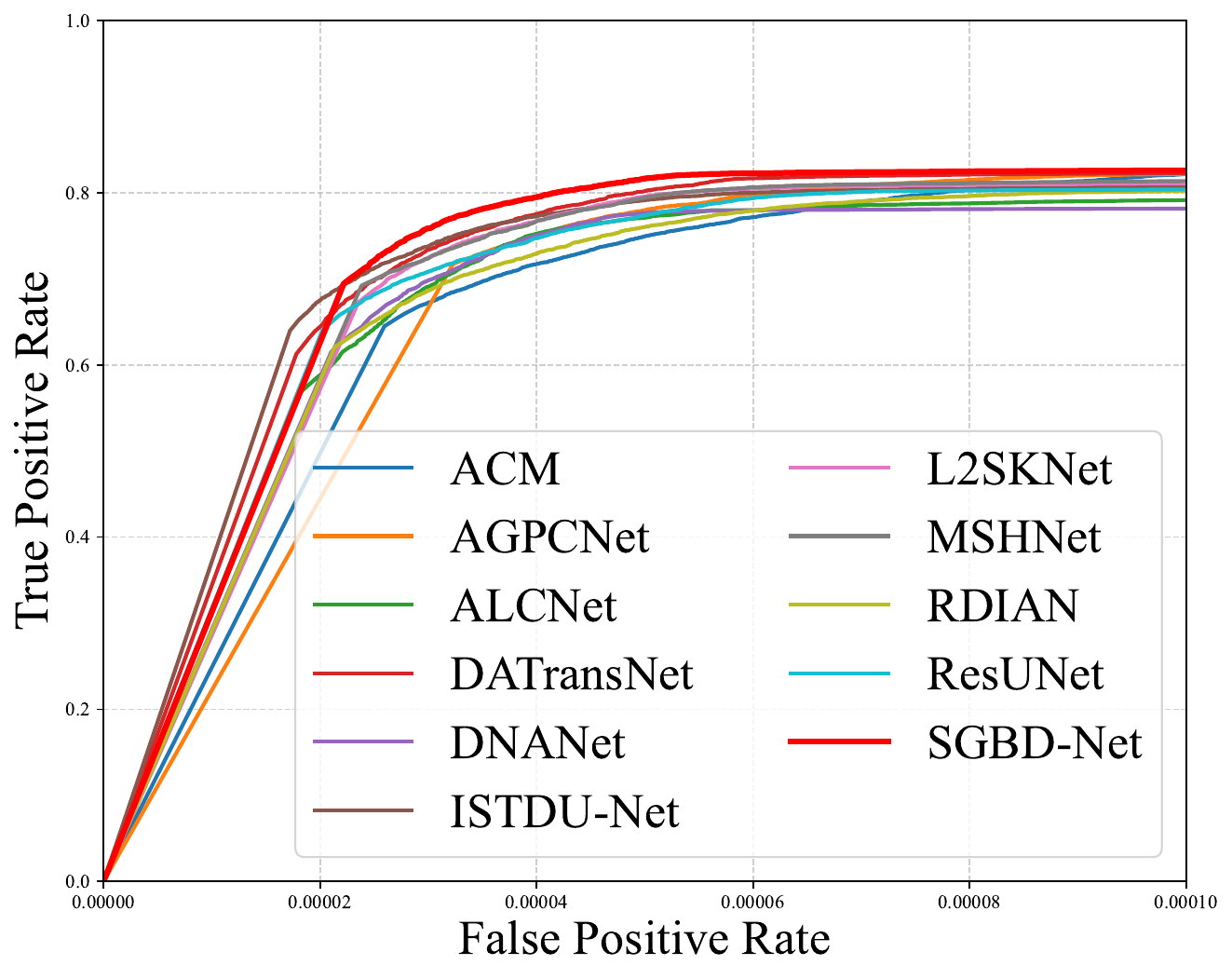}}%
    \caption{ROC curves and zoomed-in views of different single-frame networks on SIRSTD datasets.}
    \label{fig:roc_curves_1}
\end{figure*}
\subsection{Ablation Studies}

\subsubsection{Effectiveness of Orthogonality and Normalization}
We conducted an ablation study to evaluate the necessity of maintaining orthogonality and normalization, with results summarized in Table \ref{tab:orthogonality_ablation}. The orthogonal basis features are generated by applying Singular Value Decomposition (SVD) to the original feature set. Our experiments demonstrate that enforcing orthogonality does not yield measurable improvements in accuracy; instead, it introduces significant computational overhead. In contrast, normalization proved to be a critical component of our method. Consequently, we retain the normalization step while relaxing the orthogonality constraint.
Therefore, we only keep the basis features normalized.

\subsubsection{Effectiveness of Dilation Ratios in SFEM}
As described in Section \ref{SDDM_Section}, we choose to utilize the various dilation ratios to gradient information from different targets.
To determine the optimal receptive field settings for the SFEM, we conducted an ablation study evaluating various dilation ratio configurations, as detailed in Table \ref{tab:dilation_ablation}. Experimental results demonstrate that the [1, 2, 3] configuration significantly outperforms other settings. 
Consequently, we adopt this dilation configuration for our model.

\begin{figure}
    \centering
    \includegraphics[width=\linewidth]{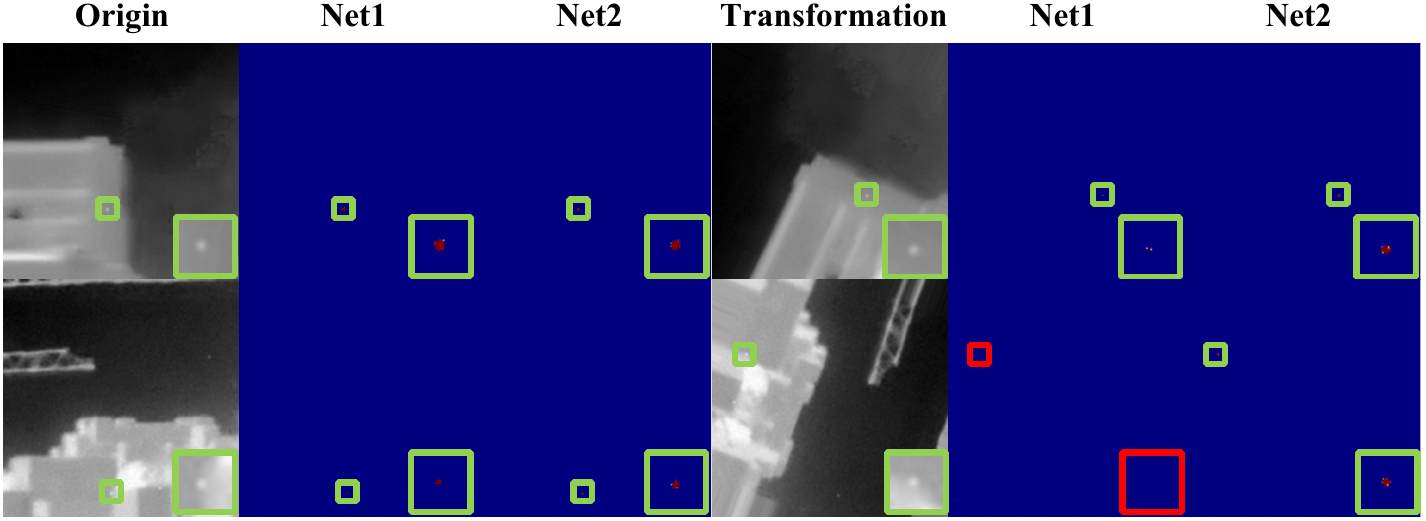}
    \caption{Comparison of results between the original and transformed images. Net1 and Net2 denote the proposed SGBD-Net and its variant without the GDM, respectively.}
    \label{compare_with_change}
\end{figure}
\begin{figure*}
    \centering
    \includegraphics[width=\linewidth]{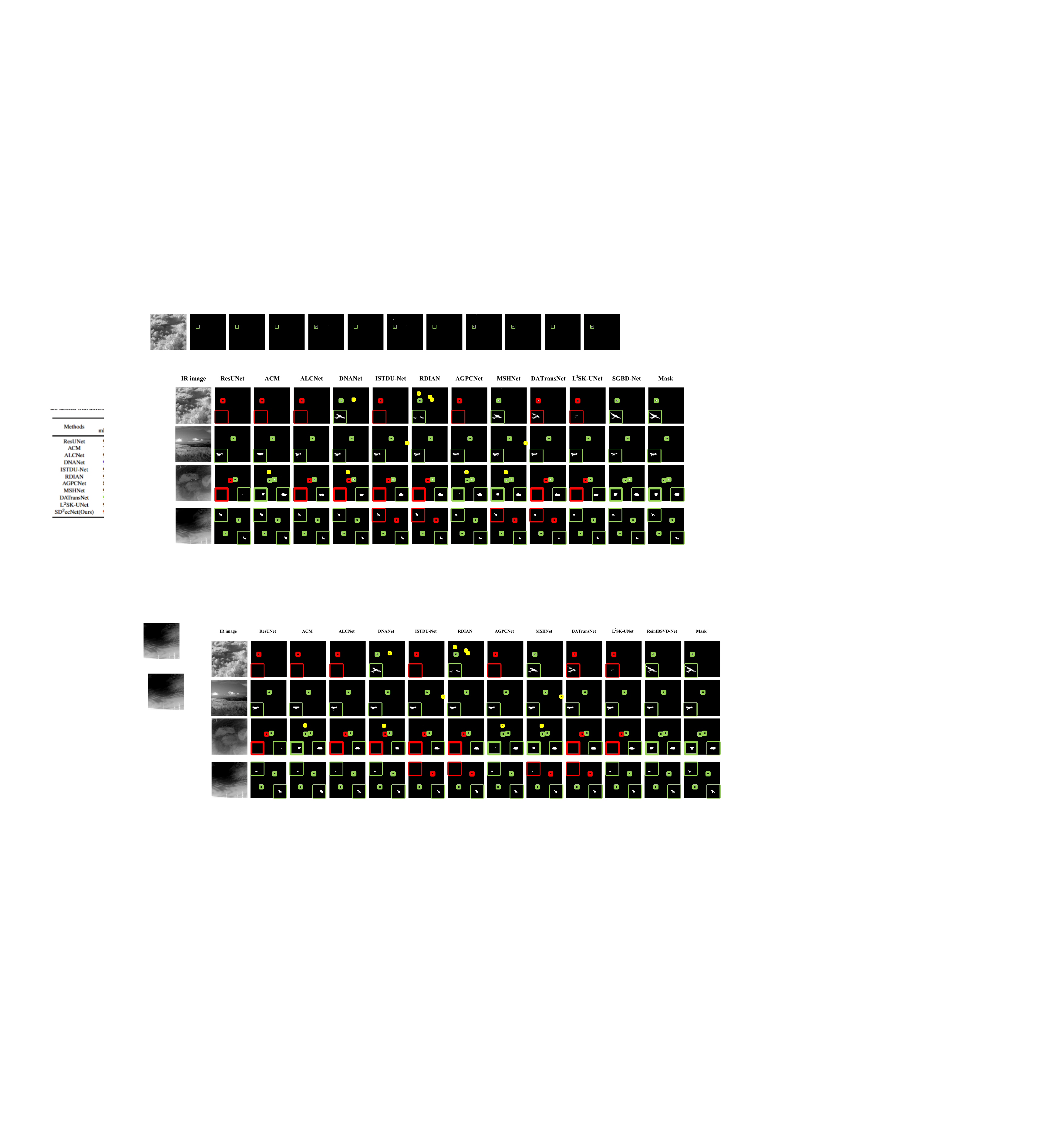}
    \caption{Qualitative results of different single-frame networks. Green, yellow, and red boxes denote correctly detected targets, missed targets, and false alarms, respectively.}
    \label{fig:visualcomparsion}
\end{figure*}
\begin{figure}
    \centering
    \includegraphics[width=.8\linewidth]{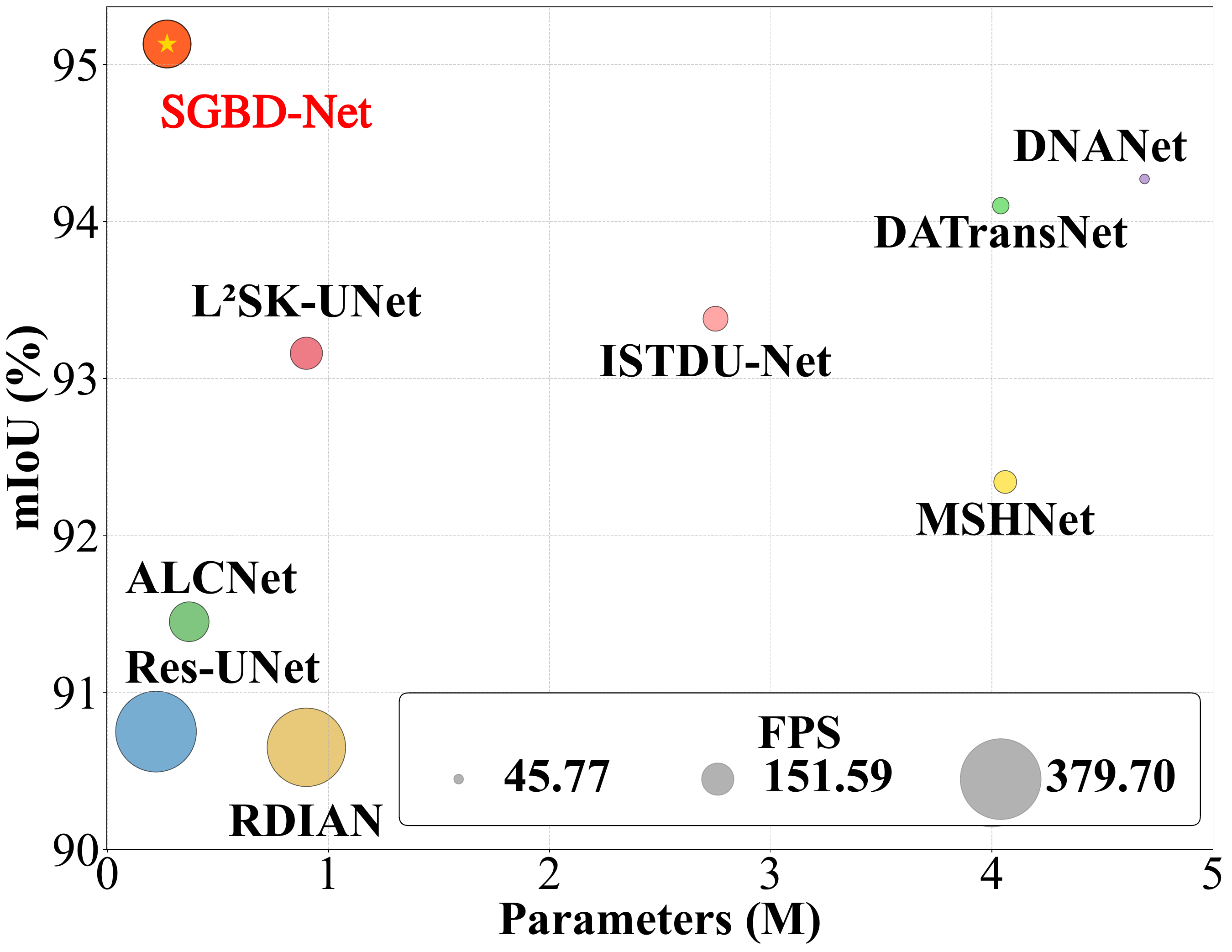}
    \caption{Comparative analysis of mIoU, parameter efficiency, and inference speed across various single-frame networks on NUDT-SIRST.}
    \label{fig:FPS_mIoU}
\end{figure}
\begin{figure}
    \centering
    \includegraphics[width=\linewidth]{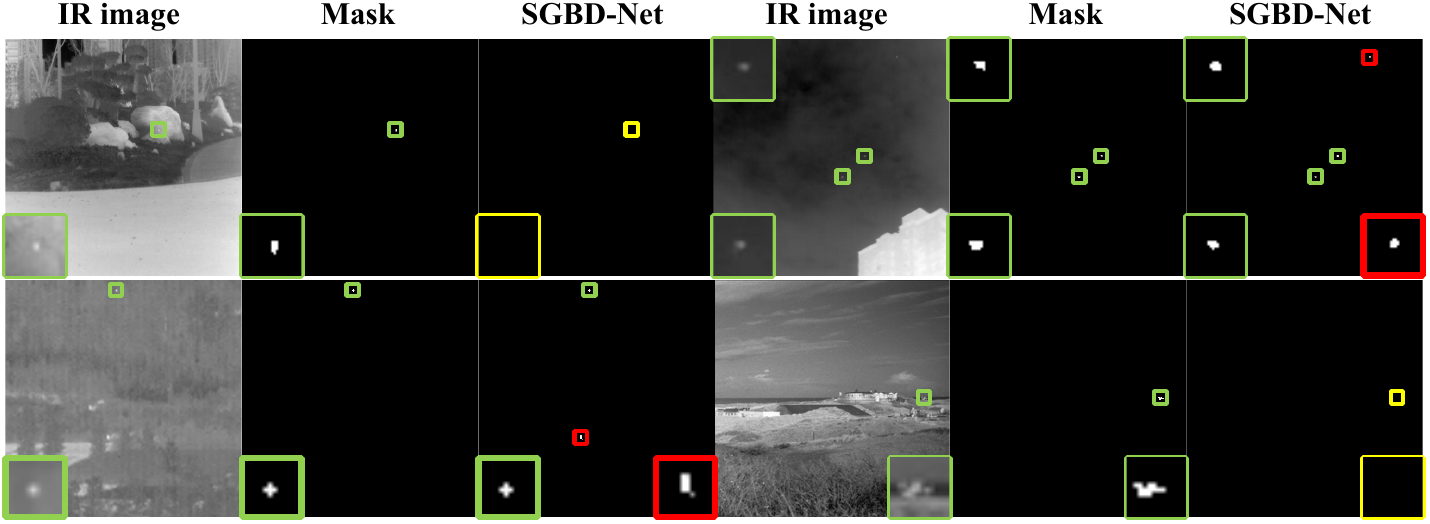}
    \caption{The failure cases of SGBD-Net. Green, yellow, and red boxes
denote correctly detected targets, missed targets, and false alarms, respectively.}
    \label{fig_failure_of_SGBD_Net}
\end{figure}
\begin{figure*}
    \centering
    \includegraphics[width=0.8\linewidth]{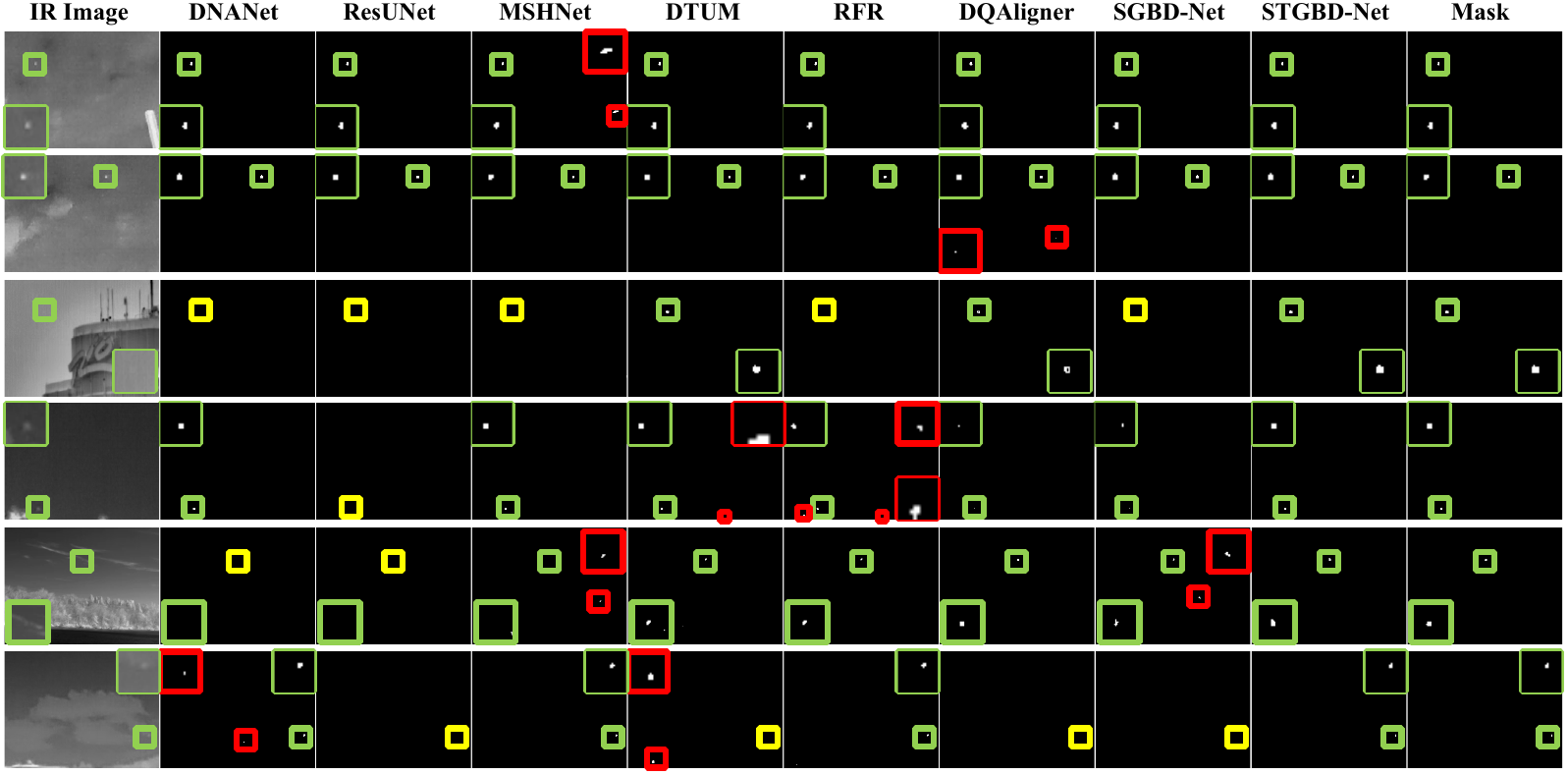}
    \caption{Qualitative results of single-frame and multi-frame networks on MIRSTD datasets. Green, yellow, and red boxes denote correctly detected targets, missed targets, and false alarms, respectively.}
    \label{fig:comparative_result_in_MIRSTD}
\end{figure*}
\begin{table*}
\centering
\caption{Comparison of single-frame detection quantitative results of different single-frame networks on SIRSTD datasets. We bold the best results and underline the second best.}
\begin{adjustbox}{max width=\linewidth}
\begin{threeparttable}
\setlength{\tabcolsep}{3pt}
\renewcommand{\arraystretch}{1.15}


\begin{tabular}{c||c|cccc|cccc|ccc}
\toprule[1.5pt]
\multirow{2}{*}{\bfseries Methods} 
&\multirow{2}{*}{\bfseries Venue} 
&\multicolumn{4}{c|}{\bfseries NUDT\mbox{-}SIRST } 
&\multicolumn{4}{c|}{\bfseries IRSTD\mbox{-}1K} 
&\bfseries Params $\downarrow$
&\bfseries FPS $\uparrow$
&\bfseries GFLOPs$\downarrow$\\
&\MetricHead \MetricHead &(M) &(frames/s) &(10\textsuperscript{9}\,FLOPS)\\
\midrule
\midrule
\rowcolor{headbg}\multicolumn{13}{l}{\textcolor{gray}{\textit{UNet-Based Methods}}}\\
Res-UNet \cite{xiao2018resUNet}& ITME\textsuperscript{18}&90.75 & 96.50 &8.30 & 96.44 & 65.50 & 89.56 &14.75 &83.66 &  \textbf{0.22} & \textbf{379.70} &0.98\\
\rowcolor{headbg}ACM\cite{DaiWuZhouBarnard2021Asymmetric}& WACV\textsuperscript{21}& 70.40 & 95.03 & 16.55 & 95.03  & 63.34 & 90.90 & 21.35 &88.39 & 0.40 & 281.55 &\underline{0.40} \\
ALCNet\cite{dai2021attentional}& WACV\textsuperscript{21}& 91.45 & 97.98 & 5.68 & 97.88 & 65.72 & 88.21 & 23.59 & 83.79 & 0.37 & 186.83 &3.74\\
\rowcolor{headbg}
DNANet\cite{RenLiHanShu2021DNANet}&TIP\textsuperscript{22} &\underline{94.27}&98.20 &5.42 & 97.44 &65.74 &90.91 &\textbf{10.48} &80.38 & 4.69 & 45.77 &14.26 \\
ISTDU-Net\cite{HouZhangTanXiZhengLi2022ISTDU-Net}&GRSL\textsuperscript{22} &93.38 &98.30 &4.87 &98.91& \underline{68.50} & 90.40 & \underline{10.82} & 88.16 & 2.75 & 117.27 &7.94\\
\rowcolor{headbg}
FC3-Net\cite{fc3net2022mingjinzhang}&ACMM\textsuperscript{22} &81.70 &96.08 &14.48 &96.62 &62.27 &89.22 &15.64 &88.41 &6.89 &205.44 & 2.63\\
RDIAN \cite{SunBaiYangBai2023Receptive-Field}&TGRS\textsuperscript{23} &90.65 & 98.41 & 13.83 & 98.76 & 64.23 & 90.11& 20.25 &88.43 & 0.90 & \underline{368.97}  &3.71\\
\rowcolor{headbg}
AGPCNet\cite{ZhangCaoPuPeng2021AGPCNet}& TAES\textsuperscript{23}&85.79 & 97.14 & 7.17 & 97.18 & 66.00 & 90.23 & 21.14 & 85.99 & 12.36 & 43.18 &12.36\\
SCTransNet\cite{Yuan2024SCtransNet}&TGRS\textsuperscript{24}&94.04 &97.78 &\underline{2.98} &98.63 &65.25 &91.91 &11.20 &87.98 &11.19 & 89.19 &10.11\\
\rowcolor{headbg}
MSHNet\cite{Liu2024MSHNet}&CVPR\textsuperscript{24}&92.34 &\textbf{98.52} &9.33 &98.69 &67.10 &\textbf{92.59}
 & 16.72 &85.90 & 4.06 & 107.44 &6.10\\
DATransNet\cite{hu2025datransnet}&GRSL\textsuperscript{25}&94.10 &98.20 &5.70 & 98.51 &67.86 &89.22 & 17.40 & 85.41 & 4.04 & 77.83 &9.19\\
\rowcolor{headbg}
L${^\mathrm{2}}$SK-UNet\cite{Wu2025Salm}&TGRS\textsuperscript{25} & 93.16 &97.14 &5.97 & 98.16 &66.92 &\underline{92.25} &16.36 &84.55 & 0.90 & 151.59 &6.89 \\
HDNet\cite{Xu2025HDNet} & TGRS\textsuperscript{25} &94.00 &97.98 &3.47 &98.61 &68.43 &90.23 &24.52 &86.43&3.67 &196.32 &5.67\\
\midrule
\rowcolor{headbg}\multicolumn{13}{l}{\textcolor{gray}{\textit{Deep Unfolding-Based Methods}}}\\
RPCANet\cite{wu2024rpcanet}&WACV\textsuperscript{24}&89.25 &96.93 &24.66 &98.38 &64.71 &89.56 &19.04 &86.93 &0.68 &174.89 &44.56\\
\rowcolor{headbg}
RPCANet++\cite{wu2025rpcanet_pp}&arXiv\textsuperscript{25}&89.33 &\underline{98.51} &6.80 &\textbf{99.36} &65.99 &88.55 &11.52 &88.38 &2.91 & 49.71 &190.67 \\
L-RPCANet\cite{liu2025lrpcanet}&arXiv\textsuperscript{25}&92.80 &97.98 &4.71 &98.48 &65.71 &88.55 &25.36 &88.49 &0.48 &120.60 & 30.88\\
\midrule
\midrule
\rowcolor{headbg}\multicolumn{13}{l}{\textcolor{gray}{\textit{SAM-Based Methods}}}\\
IRSAM\cite{zhang2024IRSAM}&ECCV\textsuperscript{24}&79.23 &96.61 &25.09 &95.53 &59.31 &87.87 &28.90 &84.69 &10.04 &131.78 &11.34 \\
\midrule
\rowcolor{headbg}
\multicolumn{13}{l}{\textcolor{gray}{\textit{Our Methods}}}\\
\addlinespace[1pt]
SGBD-Net (Ours)& -- &\textbf{95.13} &97.88 &\textbf{2.50} &\underline{98.96} & \textbf{69.40} &91.26 &13.63 &\textbf{89.68} & \underline{0.27} &217.06 &1.51  \\
\toprule[1.5pt]
\end{tabular}

\end{threeparttable}
\end{adjustbox}
\label{SIRSTD}
\end{table*}

\subsubsection{Effectiveness of the Number of Channels in the Three-stage U-Net}
To determine the optimal channel dimensions, we evaluated various configurations, as detailed in Table~\ref{tab:channel_ablation}. Scaling the channel capacities from [4, 8, 16, 32] up to [32, 64, 128, 256] yielded consistent improvements in mean Intersection over Union (mIoU). However, these performance gains were accompanied by a substantial increase in computational overhead. We found that the [8, 16, 32, 64] configuration achieved the most favorable balance between segmentation accuracy and computational efficiency; therefore, we adopted it for our final architecture.

\subsubsection{Effectiveness of the GDM}
To evaluate the effectiveness and uniqueness of the GDM-based structure, we conduct a series of comparative experiments, where we replace the SGDM with other modules. 
At first, we compare our GDM-based model with other decomposition-based modules on the SIRSTD dataset, as summarized in Table~\ref{Quantitative Module Comparison}. 
In addition, we also introduce the comparative experiments between our module with established dynamic convolution methods on the NUDT-SIRST dataset to demonstrate our advantages compared to the self-attention mechanism, as shown in Table \ref{Quantitative dynamic convolution Comparison}.
Using FPS as our efficiency metric, our model outperforms others, demonstrating a lower computational burden for GDM.

\subsubsection{Effectiveness of Core Modules in SGBD-Net}
We conduct ablation studies to validate the effectiveness of each module in our framework.
The experimental results are summarized in Table~\ref{table_ablation_sdecnet}. All experiments are conducted on two SIRSTD datasets: NUDT-SIRST and IRSTD-1K. 
Using only the SGDM module achieves competitive results on NUDT-SIRST, confirming its effectiveness for target feature extraction. Additionally, the InceptionPooling module and SGDDM module also enhance the network's performance. 
Integrating all three modules achieves the optimal balance of performance.
These results validate the effectiveness of all proposed modules.

\subsubsection{Robustness of GDM}
To demonstrate the robustness of the GDM, we compare the proposed SGBD-Net against the variant where the GDM is replaced by standard convolution layers, as illustrated in Figure \ref{compare_with_change}. For this evaluation, the original images are rotated to generate the transformed inputs.
When subjected to rotation, the baseline model often fails to detect targets in the transformed images, despite performing well on the original inputs. 
However, the model with GDM could detect the targets precisely. 
The main reason is the weights for different directions are fixed when we use convolutions, while it is dynamic when it comes to GDM. 
This demonstrates that introducing the GDM module significantly improves the network's robustness.

\label{sec:experiment}
\subsection{Comparison with state-of-the-art (SOTA) Methods in SIRSTD}

\subsubsection{Qualitative Results Comparison}
As illustrated by the Receiver Operating Characteristic (ROC) curves in Fig. \ref{fig:roc_curves_1}, our model's curve converges rapidly toward the top-left coordinate, outstripping contemporary deep learning baselines. 
This consistent margin across a wide range of operational thresholds validates the robustness and discriminative performance of our network in diverse infrared environments.
Additionally, qualitative results in Fig. \ref{fig:visualcomparsion} underscore the performance disparity between our architecture and existing benchmarks. 
While our network preserves fine-grained target geometry, most mainstream methods exhibit significant limitations in maintaining structural integrity. 
For instance, RDIAN \cite{SunBaiYangBai2023Receptive-Field} struggles to capture localized details, often resulting in fragmented or incomplete segmentations. 
Furthermore, while many SIRSTD methods prioritize false alarm suppression, they frequently suffer from missed detections in the presence of cluttered backgrounds and low-contrast targets. 
In contrast, SGBD-Net achieves precise localization with a negligible false alarm rate.

\subsubsection{Quantitative Result Comparison}
We compare our network with various categories of networks, including UNet-based, Deep Unfolding-Based, and SAM-based methods, as shown in Table~\ref{SIRSTD}.
Extensive quantitative results demonstrate that our proposed method consistently outperforms existing SOTA approaches across multiple metrics.
Although our model contains only 0.27M parameters and involves fewer iterations compared with computationally expensive iterative modules, such as RPCANet, it still achieves superior performance, as shown in Fig. \ref{fig:FPS_mIoU}. 
While conventional SIRSTD methods mostly rely on increasing architectural depth or complexity to enhance performance, our approach achieves a significantly more favorable accuracy-efficiency trade-off. 
Specifically, our network attains 95.13\% and 69.40\% mIoU on the SIRSTD benchmarks, respectively, while requiring a mere 1.51 GFLOPS. In contrast, the prevalent DNANet demands a nearly tenfold increase in computational budget (14.46 GFLOPS). 
Rather than relying on excessive parameterization to approximate contrast-sensitive kernels, we explicitly introduce gradient-based structural priors into the network’s foundation. This design choice offloads the burden of discovering local variation patterns from the learning process. 
Consequently, the network preserves significant representational power for sub-pixel targets while maintaining computational efficiency.
Additionally, the other gradient-based algorithms, like RDIAN \cite{SunBaiYangBai2023Receptive-Field} and L\textsuperscript{2}SK-UNet \cite{Wu2025Salm}, are also inferior to our method. 
This demonstrates that our basis-decomposition-driven approach offers greater adaptive representational capacity than conventional static gradient kernels. 
Unlike RPCA-like models which perform iterative operations using the same weights, our backbone utilizes a three-stage structure and can adaptively refine target features at different stages with different weights, leading to our superior performance.

\begin{table*}
\caption{Comparison of multi-frame detection quantitative results of different multi-frame networks on MIRSTD datasets.
We bold the best results and underline the second best.}
\centering
\begin{adjustbox}{max width=\linewidth}
\begin{threeparttable}
\setlength{\tabcolsep}{3pt}
\renewcommand{\arraystretch}{1.15}


\begin{tabular}{c||c|cccc|cccc|ccc}
\toprule[1.5pt]
\multirow{2}{*}{\bfseries Methods} 
&\multirow{2}{*}{\bfseries Venue} 
&\multicolumn{4}{c|}{\bfseries NUDT-MIRSDT} 
&\multicolumn{4}{c|}{\bfseries IRDST} 
&\bfseries Params $\downarrow$
&\bfseries FPS $\uparrow$
&\bfseries GFLOPs$\downarrow$\\
&\MetricHead \MetricHead &(M) &(frames/s) &(10\textsuperscript{9}\,FLOPS)\\
\midrule
\midrule

\rowcolor{headbg}\multicolumn{13}{l}{\textcolor{gray}{\textit{Single-Frame Methods}}}\\
Res-UNet \cite{xiao2018resUNet}& ITME\textsuperscript{18}
        &58.66 &60.79 &17.90 &97.93
        &56.72 &95.56 &29.46 &98.88 &  \textbf{0.22} & \textbf{379.70} &\textbf{0.98}\\
         
\rowcolor{headbg}
ISTDU-Net\cite{HouZhangTanXiZhengLi2022ISTDU-Net}&GRSL\textsuperscript{22}
          &55.34 &67.50 &93.28 &96.09 
          &55.05&98.43&38.17&\underline{99.51} & 2.75 & 117.27 &7.94\\

DNANet \cite{RenLiHanShu2021DNANet}&TIP\textsuperscript{22}
       &57.94&70.79&92.16&98.84 
       &57.83&\underline{99.48}&31.73&99.39 & 4.69 & 45.77 &14.26\\
\rowcolor{headbg}
MSHNet\cite{Liu2024MSHNet}&CVPR\textsuperscript{24}
        &50.99&64.14&103.81&94.50 
        &57.58&\textbf{99.74}&36.69&99.23& 4.06 & 107.44 &6.10\\

L${^\mathrm{2}}$SK-UNet\cite{Wu2025Salm}&TGRS\textsuperscript{25}
&44.87&63.97&130.24&92.45
&56.82&96.34&26.78&98.88 & 0.90 & 151.59 &6.89

\\

\midrule

\rowcolor{headbg}\multicolumn{13}{l}{\textcolor{gray}{\textit{Multi-Frame Methods}}}\\
Res-UNet+DTUM\cite{Li2025dtum}& TNNLS\textsuperscript{23} &52.04 &62.70 &72.83 &94.09 & 57.18 &97.65 &37.53 &99.46& 0.30 & 86.91 & 10.30 
 \\
\rowcolor{headbg}
ISTDU-Net+DTUM\cite{Li2025dtum}& TNNLS\textsuperscript{23} &85.84 &\textbf{97.80} &1.10 &99.53 &56.62 &98.43&31.02 &99.41& 2.82 & 21.34 & 45.18  \\
DNANet+DTUM\cite{Li2025dtum}& TNNLS\textsuperscript{23} &84.15 &94.39 &1.57 &98.66 
                &\underline{58.88} &98.95 &30.18 &99.33 & 4.77 & 10.37 & 76.75 \\
\rowcolor{headbg}
MSHNet+DTUM\cite{Li2025dtum}& TNNLS\textsuperscript{23} &85.57 &\underline{96.70} &1.53 &99.15 
              &56.57 &98.82 &27.98 &98.70  & 4.14 & 22.88 & 35.98 \\
L${^\mathrm{2}}$SK-UNet+DTUM\cite{Li2025dtum}& TNNLS\textsuperscript{23} &\underline{86.21} &96.12 &\textbf{0.27} &99.08
             &53.52 &98.04 &43.69 &99.07  & 0.97 & 38.91 & 39.93 \\
Res-UNet+RFR\cite{ying2025rfr}& TGRS\textsuperscript{25} &85.74 &96.41 &2.19 &\underline{99.40} 
               &56.70 &94.77 &\textbf{24.78 }&98.47& 0.33 & 60.06 & 20.17  \\
\rowcolor{headbg}
ISTDU-Net+RFR\cite{ying2025rfr}& TGRS\textsuperscript{25} &61.61 &72.53 &78.94 &95.21 
              & 58.24 & 98.82 & \underline{27.14} & 99.49& 2.86 & 19.06 & 55.30  \\
DNANet+RFR\cite{ying2025rfr}&TGRS\textsuperscript{25} &82.43 &86.99 &3.02 &96.95 
                & 56.88 & 97.78 & 27.42 & 99.03 & 4.80 & 10.81 & 86.83 \\
\rowcolor{headbg}
MSHNet+RFR\cite{ying2025rfr}& TGRS\textsuperscript{25} & 75.65 & 79.99 & 20.11 &98.15
              &56.60 &96.60 &23.45 &99.33 & 4.17 & 24.21 & 45.46\\
L${^\mathrm{2}}$SK-UNet+RFR\cite{ying2025rfr}& TGRS\textsuperscript{25} &82.82 &86.41 &2.06 &99.39 
             & 55.41 & 97.52 & 43.73 & 99.35 & 1.01 & 33.45 & 50.04\\
\rowcolor{headbg}
DQAligner\cite{Deng2026DQAligner}& TGRS\textsuperscript{26} &85.73 &91.56 &1.25 &\textbf{99.56} 
              &56.99 & 96.34 & 26.46 & 98.59 & 0.55 & 26.23 & 27.56 \\
\midrule

\rowcolor{headbg}\multicolumn{13}{l}{\textcolor{gray}{\textit{Our Single-Frame Methods}}}\\
SGBD-Net (Ours)  &-
&61.92 &71.78 &84.08 &94.06 
&56.34 &96.73 &37.01 &98.58 &\underline{0.27} &\underline{217.06} &\underline{1.51}\\

\midrule
\rowcolor{headbg}\multicolumn{13}{l}{\textcolor{gray}{\textit{Our Multi-Frame Methods}}}\\
STGBD-Net (Ours) &-
&\textbf{87.68} &96.59 &\underline{0.79} &99.19 
&\textbf{59.26} &99.08 &31.18 &\textbf{99.56} 
& 0.28 & 33.88 & 8.32  \\
\toprule[1.5pt]
\end{tabular}

\end{threeparttable}
\end{adjustbox}
\label{MIRSTD_table_ablation_stdecnet}
\end{table*}

\begin{figure}
    \centering
    \includegraphics[width=\linewidth]{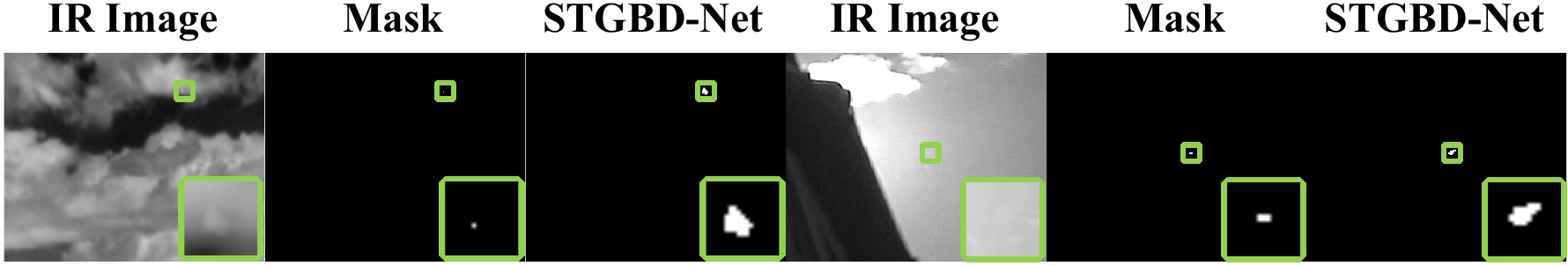}
    \caption{The failure cases of STGBD-Net, where all predicted targets are enlarged.}
    \label{failure_result_in_MIRSTD}
\end{figure}

\subsubsection{Failure Cases and Limitations}
Despite achieving state-of-the-art performance, SGBD-Net still encounters challenges in extreme scenarios where targets and background noise share nearly identical structural characteristics, as shown in Fig. \ref{fig_failure_of_SGBD_Net}. In these instances, the feature representations are so indistinguishable that differentiating targets from false alarms becomes difficult even for a human observer.
To resolve this ambiguity, it becomes essential to incorporate temporal information.

\subsection{Comparison with SOTA Methods in MIRSTD}
\subsubsection{Qualitative Results Comparison}
We evaluate our models through qualitative comparisons, as illustrated in Fig. \ref{fig:comparative_result_in_MIRSTD}. The visual results demonstrate that the proposed SGBD-Net and STGBD-Net exhibit superior detection capabilities compared to other SOTA models.
Single-frame methods often suffer from false alarms and missed detections, particularly when background clutter shares similar structural characteristics with the targets.
Furthermore, we compare our model against the most recent SOTA MIRSTD methods, including RFR \cite{ying2025rfr}, DTUM \cite{Li2025dtum}, and DQAligner \cite{Deng2026DQAligner}. As observed in the visual comparisons, our proposed model consistently exhibits superior detection performance.

\subsubsection{Quantitative Result Comparison}
We further conduct a series of experiments on the NUDT-MIRSDT and SIRSTD datasets to demonstrate the superiority of our model, as shown in Table \ref{MIRSTD_table_ablation_stdecnet}. 
Notably, the NUDT-MIRSDT dataset presents a significantly more complex background than IRDST, allowing us to comprehensively evaluate our model's robustness across varying levels of difficulty.
The results clearly demonstrate that incorporating temporal information improves the network's performance. 
Furthermore, this performance gain is substantially more pronounced on the NUDT-MIRSDT dataset compared to SIRSTD, indicating that temporal information effectively suppresses the influence of severe clutter, whereas its impact is naturally more modest in simpler background conditions.
Among all single-frame approaches, SGBD-Net yields the best results and achieves a highly favorable performance-efficiency trade-off.
By introducing the TGDM to the network, we effectively upgrade SGBD-Net to STGBD-Net, enabling the model to capture critical motion cues across adjacent frames. 
This enhancement significantly boosts the mIoU from 61.92\% to 87.68\% on NUDT-MIRSDT, and from 56.34\% to 59.26\% on IRDST.
Since the only architectural difference between SGBD-Net and STGBD-Net is the inclusion of the temporal module, the performance comparison between these two models inherently serves as the ablation study. Therefore, we did not include a separate ablation experiment for this module to avoid redundancy. 
Overall, STGBD-Net establishes a new SOTA in the mIoU metric and demonstrates highly competitive performance across other key metrics.

\subsubsection{Failure Cases and Limitations}
Despite its outstanding performance, STGBD-Net still exhibits certain limitations, as illustrated in Figure \ref{failure_result_in_MIRSTD}. While the model successfully detects the targets, it expands their boundaries to include nearby background pixels, resulting in degraded localization accuracy.
A potential reason of our current framework lies in its fixed strategy for basis feature extraction.
For example, when targets are extremely small, the extraction methods should concentrate on the local neighborhood.

\section{Conclusion}
\label{sec:conclusion}
This work proposes the BDM as an extensible module and a family of networks that are built upon BDMs.  
We introduce the SGBD-Net by integrating SGDM and SGDDM into a three-stage U-Net structure for SIRSTD. 
Moreover, we develop TGDM to extract temporal features across frames and construct STGBD-Net for MIRSTD. Experiments on the public MIRSTD and SIRSTD datasets demonstrate that our networks can outperform other methods.

A potential limitation of our current framework lies in the fixed strategy for basis feature extraction, which may limit its optimal adaptability across highly diverse target morphologies. In this paper, our model enhances detailed information through a gradient-based basis feature extraction method. To address this limitation, we plan to introduce a more adaptive strategy for extracting basis features in future work.
Firstly we would enlarge our basis feature extraction methods, not only for small target but larger target. 
Then, the gating mechanism would be applied to select the basis features according to the input images and text prompts. 
Finally, our module could extract the basis features based on our specific needs. 

Furthermore, we also plan to explore applying BDM to other tasks. While this work focuses on IRSTD, the principle of task-oriented basis decomposition can be extended to other application domains, like salient object detection and drone vision tasks, by designing the corresponding BFEM and FEM. 

\bibliographystyle{IEEEtran}   
\bibliography{ref}
\begin{IEEEbiography}
[{\includegraphics[width=1in,height=1.2in,clip,keepaspectratio]{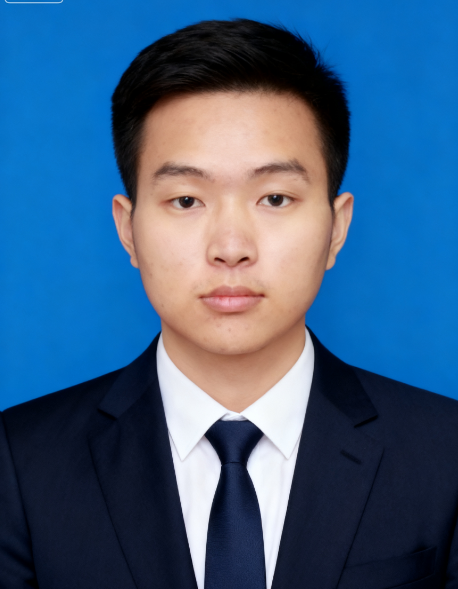}}]
{Chen Hu} (Graduate Student Member, IEEE) received the B.S. and M.E. degree from the School of Information and Communication Engineering, University of Electronic Science and Technology of China (UESTC), Chengdu, China, in 2022. Now, He is pursuing the Ph.D.  degree with the School of Intelligent Systems Engineering, Sun Yat-sen University.

His research interests include image processing, computer vision, object detection, and recognition.

\end{IEEEbiography}
\vspace{-1.2cm}
\begin{IEEEbiography}
[{\includegraphics[width=1in,height=1.25in,clip,keepaspectratio]{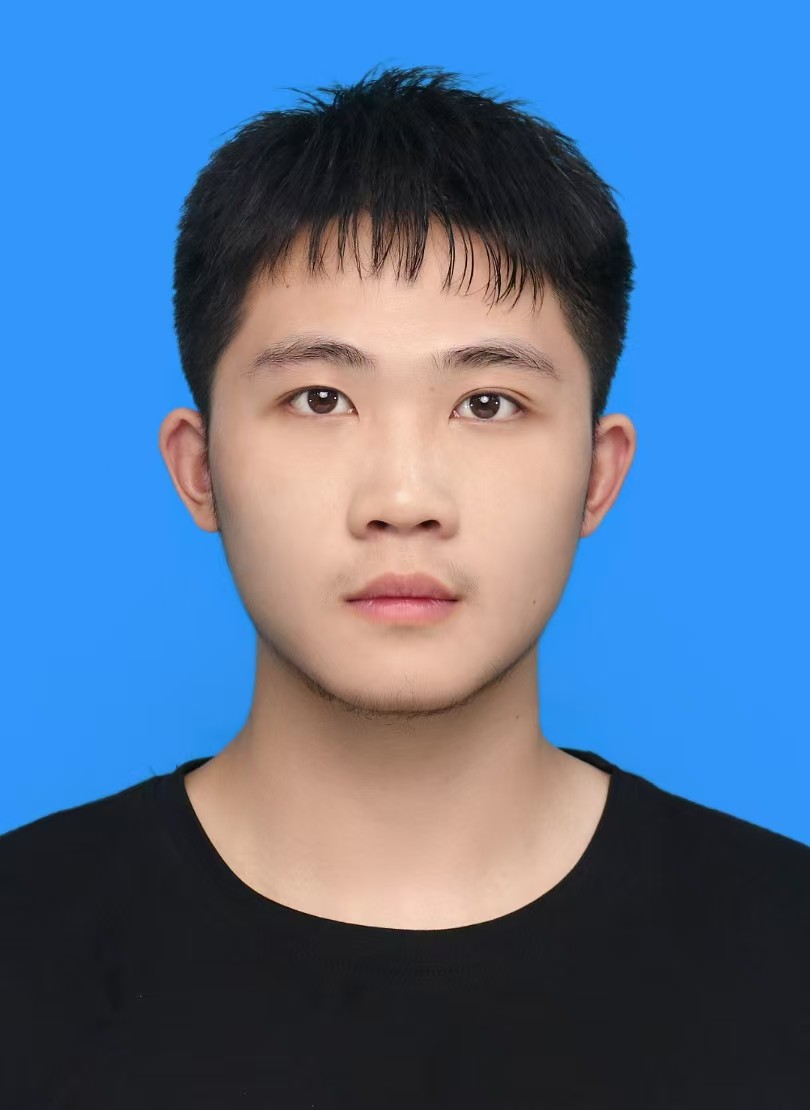}}]{Mingyu Zhou} is currently pursuing the B.E. degree in Network Engineering with the School of Information and Communication Engineering, University of Electronic Science and Technology of China (UESTC), Chengdu, China.

\end{IEEEbiography}
\vspace{-1.2cm}
\begin{IEEEbiography}[{\includegraphics[width=1in, height=1.25in, clip, keepaspectratio]{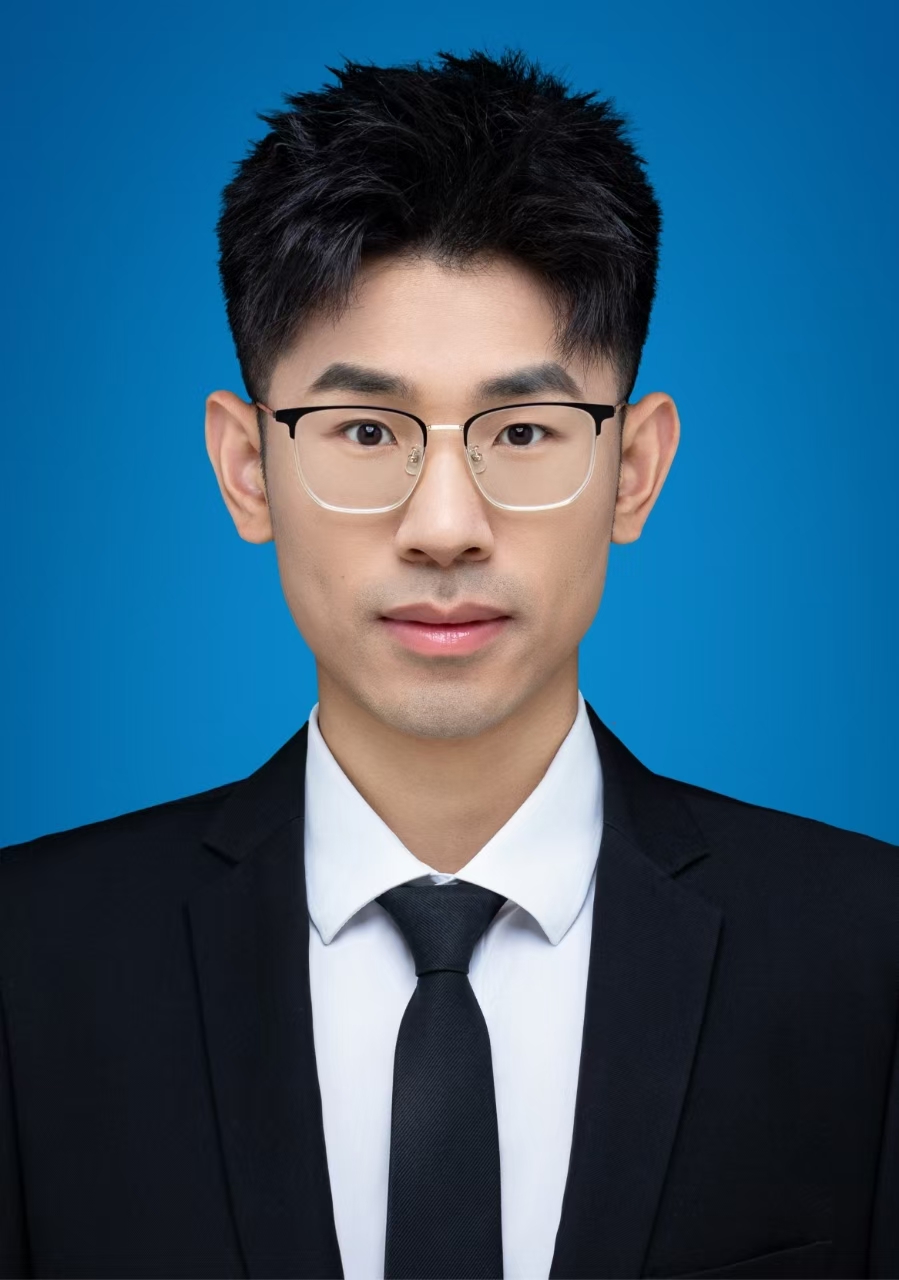}}]{Shuai Yuan} received the B.S. degree from Xi'an Technological University, Xi'an, China, in 2019, and the Ph.D. degree from Xidian University, Xi’an, China, in 2025. He was a visiting Ph.D. student at the University of Melbourne, Australia. He is currently a postdoctoral researcher with the School of Instrument Science and Opto-Electronics Engineering, Hefei University of Technology, Hefei, China, working closely with Prof. Yu Liu. His research interests include infrared image understanding, remote sensing, and deep learning. \end{IEEEbiography}
\vspace{-1.2cm}
\begin{IEEEbiography}[{\includegraphics[width=1in,height=1.25in,clip,keepaspectratio]{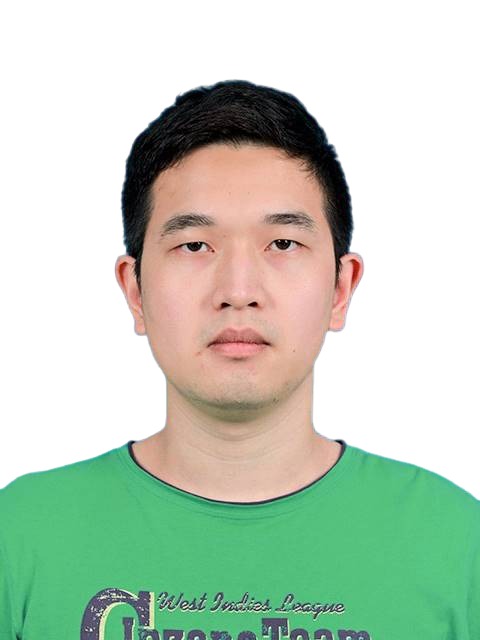}}]{Hongbo Hu}
received the B.E. degree in Optoelectronic Information Science and Engineering from University of Electronic Science and Technology of China (UESTC), Chengdu, China, in 2021 and is currently a Ph.D. student in the School of Information and Communication Engineering, UESTC. His research interests include image processing, infrared target detection, and computer vision.
\end{IEEEbiography}
\vspace{-1.2cm}
\begin{IEEEbiography}
[{\includegraphics[width=1in,height=1.25in, clip,keepaspectratio]{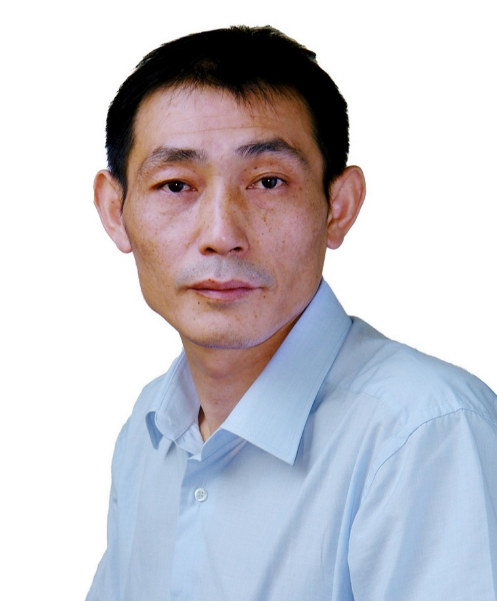}}]{Zhenming Peng} (Senior Member, IEEE) received his Ph.D. degree in geodetection and information technology from the Chengdu University of Technology, Chengdu, China, in 2001. From 2001 to 2003, he was a postdoctoral researcher at the Institute of Optics and Electronics, Chinese Academy of Sciences, Chengdu, China. He is currently a Professor with the University of Electronic Science and Technology of China, Chengdu. His research interests include image processing, machine learning, object detection, and remote sensing applications.
\end{IEEEbiography}
\vspace{-1.2cm}
\begin{IEEEbiography}
[{\includegraphics[width=1in,height=1.25in,clip,keepaspectratio]{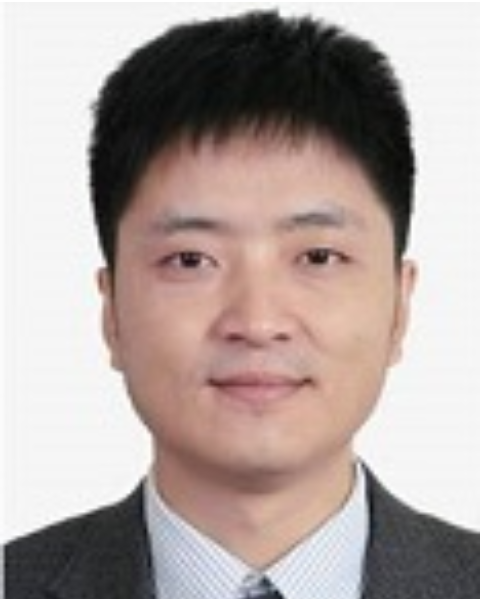}}]{Tian Pu} received the Ph.D. degree in optical engineering from the Beijing Institute of Technology, Beijing, China, in 2002.

He is currently a senior engineer at the School of Information and Communication Engineering, University of Electronic Science and Technology of China (UESTC), Chengdu, China. His research interests include image processing, infrared small target detection, and medical image analysis.
\end{IEEEbiography}
\vspace{-1.2cm}
\begin{IEEEbiography}
[{\includegraphics[width=1in,height=1.25in,clip,keepaspectratio]{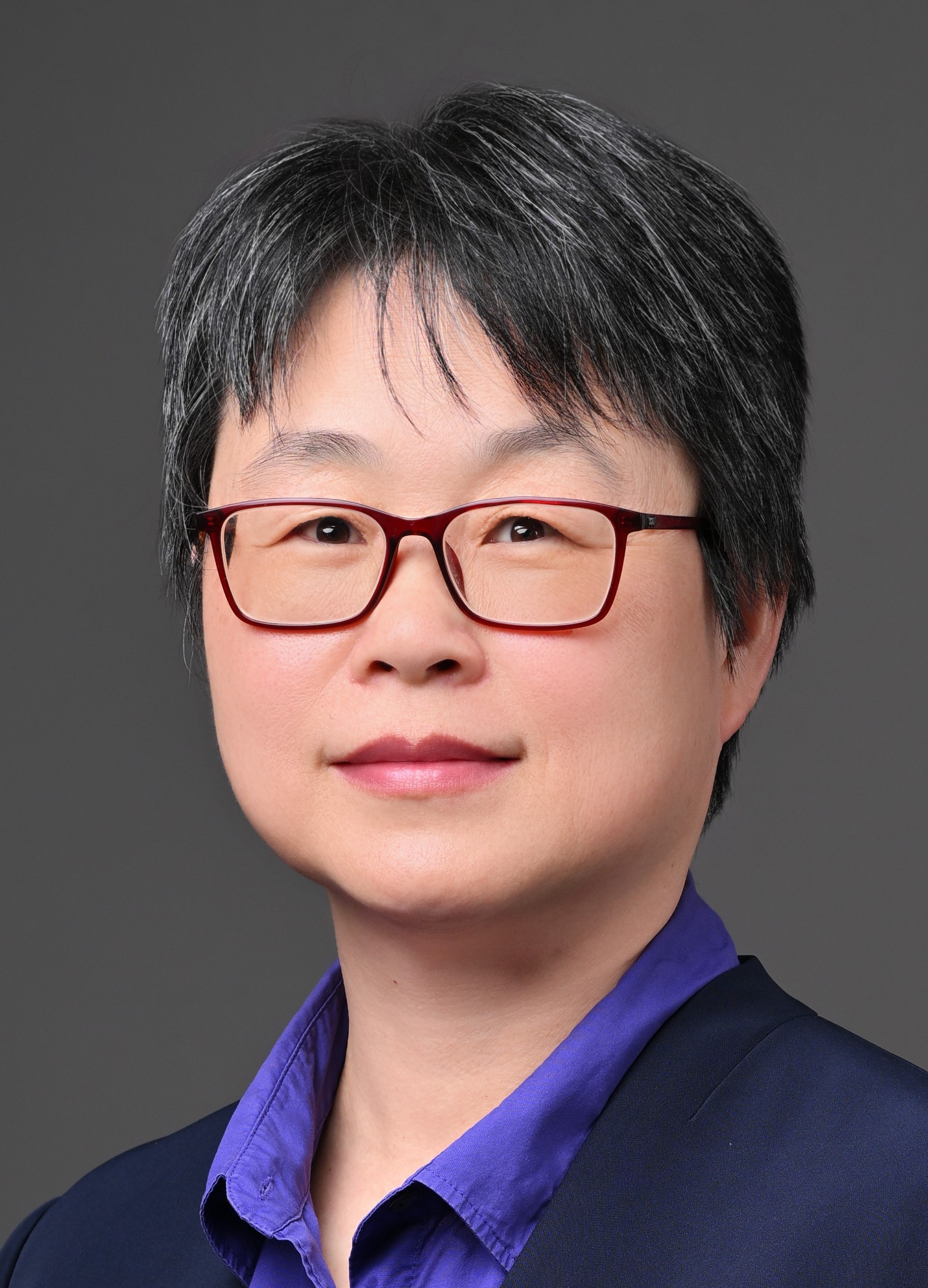}}]{Xiying Li} (Member, IEEE) received the Ph.D.
degree in optical engineering from the Beijing
Institute of Technology in 2002. She is currently
a Professor with the School of Intelligent Systems
Engineering, Sun Yat-sen University. Her research
interests include intelligent transportation systems,
traffic information collection, traffic video, and
image big data processing and application.

\end{IEEEbiography}

\end{document}